\documentclass{article}

\usepackage[final]{neurips_2019}

\usepackage[utf8]{inputenc} %
\usepackage[T1]{fontenc}    %
\usepackage{hyperref}       %
\usepackage{url}            %
\usepackage{booktabs}       %
\usepackage{amsfonts}       %
\usepackage{nicefrac}       %
\usepackage{microtype}      %
\usepackage{amsmath}
\usepackage{amssymb}
\usepackage{enumerate} 
\usepackage{graphicx}
\usepackage{subcaption}
\usepackage{algorithm}
\usepackage{algorithmic}
\usepackage{color}
\usepackage{wrapfig}
\usepackage{amsthm}
\usepackage{tikz}
\usepackage{multirow}
\usepackage{bm}
\usepackage{mathrsfs}
\usepackage{xfrac}
\usepackage{enumitem}
\usepackage[font=small,labelfont=bf]{caption}

\definecolor{mydarkblue}{rgb}{0,0.1,0.45}
\definecolor{amber}{rgb}{1.0, 0.75, 0.0}
\hypersetup{
    colorlinks=true,
    linkcolor=mydarkblue,
    citecolor=mydarkblue,
    filecolor=mydarkblue,
    urlcolor=mydarkblue}

\newtheorem{thm}{Theorem}

\newcommand{\real}{\mathbb{R}}

\newcommand{\expect}{\mathbb{E}}
\newcommand{\vars}{\mathbb{V}}

\newcommand{\loss}{\mathcal{L}}

\newcommand{\normal}{\mathcal{N}}

\newcommand{\weight}{\mathbf{W}}
\newcommand{\fisher}{\mathbf{F}}
\newcommand{\kron}{\otimes}

\newcommand{\hessian}{\mathbf{H}}
\newcommand{\transHessian}{\tilde{\hessian}}
\newcommand{\covariance}{\mathbf{C}}
\newcommand{\transCovariance}{\tilde{\covariance}}

\newcommand{\precon}{\mathbf{P}}

\newcommand{\iden}{\mathbf{I}}
\newcommand{\params}{\bm{\theta}}
\newcommand{\lr}{\alpha}

\newcommand{\cov}{\mathrm{Cov}}

\newcommand{\bS}{\mathbf{S}}
\newcommand{\bA}{\mathbf{A}}

\newcommand{\newvec}{\mathrm{vec}}
\newcommand{\ba}{\mathbf{a}}
\newcommand{\bs}{\mathbf{s}}

\newcommand{\bigO}{\mathcal{O}}

\title{Which Algorithmic Choices Matter at Which Batch Sizes?  Insights From a Noisy Quadratic Model}

\author{Guodong Zhang${}^{1, 2, 3}$\thanks{Work done as part of the Google Student Researcher Program. Email: \texttt{gdzhang@cs.toronto.edu}} , Lala Li${}^{3}$, Zachary Nado${}^{3}$, James Martens${}^{4}$, \\ \textbf{Sushant Sachdeva${}^{1}$, George E. Dahl${}^{3}$, Christopher J. Shallue${}^{3}$, Roger Grosse${}^{1, 2}$} \\ ${}^{1}$University of Toronto, ${}^{2}$Vector Institute, ${}^{3}$Google Research, Brain Team, ${}^{4}$DeepMind} %

\begin{document}

\maketitle

\begin{abstract}
Increasing the batch size is a popular way to speed up neural network training, but beyond some critical batch size, larger batch sizes yield diminishing returns. In this work, we study how the critical batch size changes based on properties of the optimization algorithm, including acceleration, preconditioning and averaging, through two different lenses: large scale experiments, and analysis of a simple noisy quadratic model (NQM). We experimentally demonstrate that optimization algorithms that employ preconditioning, specifically Adam and K-FAC, result in much larger critical batch sizes than stochastic gradient descent with momentum. We also demonstrate that the NQM captures many of the essential features of real neural network training, despite being drastically simpler to work with. The NQM predicts our results with preconditioned optimizers and exponential moving average, previous results with accelerated gradient descent, and other results around optimal learning rates and large batch training, making it a useful tool to generate testable predictions about neural network optimization.

\end{abstract}

\section{Introduction}

Increasing the batch size is one of the most appealing ways to accelerate neural network training on data parallel hardware. Larger batch sizes yield better gradient estimates and, up to a point, reduce the number of steps required for training, which reduces the training time. The importance of understanding the benefits of modern parallel hardware has motivated a lot of recent work on training neural networks with larger batch sizes~\citep{goyal2017accurate, osawa2018second, mccandlish2018empirical, shallue2018measuring}. To date, the most comprehensive empirical study of the effects of batch size on neural network training is \citet{shallue2018measuring}, who confirmed that increasing the batch size initially achieves perfect scaling (i.e.~doubling the batch size halves the number of steps needed) up to a problem-dependent critical batch size, beyond which it yields diminishing returns~\citep{balles2017coupling, goyal2017accurate, jastrzkebski2017three, mccandlish2018empirical}. \citet{shallue2018measuring} also provided experimental evidence that the critical batch size depends on the optimization algorithm, the network architecture, and the data set. However, their experiments only covered plain SGD, SGD with (heavy-ball) momentum, and SGD with Nesterov momentum, leaving open the enticing possibility that other optimizers might extend perfect scaling to even larger batch sizes. 

Empirical scaling curves like those in \citet{shallue2018measuring} are essential for understanding the effects of batch size, but generating such curves, even for a single optimizer on a single task, can be very expensive.
On the other hand, existing theoretical analyses that attempt to analytically derive critical batch sizes (e.g. \citet{ma2017power, yin2018gradient, jain2018parallelizing}) do not answer our questions about which optimizers scale the best with batch size. They tend to make strong assumptions, produce parameter-dependent results that are difficult to apply, or are restricted to plain SGD. It would be ideal to find a middle ground between a purely empirical investigation and theoretical analysis by building a model of neural network optimization problems that captures the essential behavior we see in real neural networks, while still being easy to understand. Additionally, we need to study optimizers beyond momentum SGD since they might provide us an approach to exploit speedups from the very largest batch sizes. In this work, we make the following contributions:
\begin{enumerate}[leftmargin=0.8cm] %
    \setlength\itemsep{0.05em}
    \item We show that a simple noisy quadratic model (NQM) is remarkably consistent with the batch size effects observed in real neural networks, while allowing us to run experiments in seconds, making it a great tool to generate testable predictions about neural network optimization.
    \item We show that the NQM successfully predicts that momentum should speed up training relative to plain SGD at larger batch sizes, but have no benefit at small batch sizes.
    \item Through large scale experiments with Adam~\citep{kingma2014adam} and K-FAC~\citep{martens2015optimizing}, we confirm that, as predicted by the NQM, preconditioning extends perfect batch size scaling to larger batch sizes than are possible with momentum SGD alone. Furthermore, unlike momentum, preconditioning can help at small batch sizes as well. 
    \item Lastly, we show that, as predicted by the NQM, exponential moving averages reduce the number of steps required for a specific batch size and can achieve the same acceleration with smaller batch sizes, thereby saving computation.
\end{enumerate}

\section{Related Work}

In a classic paper, \citet{bottou2008tradeoffs} studied the asymptotics of stochastic optimization algorithms and found SGD to be competitive with fancier approaches. They showed that stochastic optimization involves fundamentally different tradeoffs from full-batch optimization.
More recently, several studies have investigated the relationship between batch size and training time for neural networks. \citet{chen2018effect} studied the effect of network width on the critical batch size, and showed experimentally that it depends on both the data set and network architecture. \citet{golmant2018computational} studied how various heuristics for adjusting the learning rate as a function of batch size affect the relationship between batch size and training time. 
\citet{shallue2018measuring} conducted a comprehensive empirical study on the relationship between batch size and training time with different neural network architectures and data sets using plain SGD, heavy-ball momentum, and Nesterov momentum. Finally, \citet{mccandlish2018empirical} used the average gradient noise over training to predict the critical batch size. All of these studies described a basic relationship between batch size and training steps to a fixed error goal, which is comprised of three regions: perfect scaling initially, then diminishing returns, and finally no benefit for all batch sizes greater than the critical batch size.

Other studies have attempted to characterize the critical batch size analytically in stochastic optimization. Under varying assumptions, \citet{ma2017power, yin2018gradient, jain2018parallelizing} all derived analytical notions of critical batch size, but to our knowledge, all for SGD. 

Additionally, previous studies have shown that SGD and momentum SGD are equivalent for small learning rates (after appropriate rescaling), both for the continuous limit~\citep{leen1994} and discrete settings~\citet{yuan2016}. However, they do not explain why momentum SGD (including heavy-ball and Nesterov momentum) sometimes outperforms plain SGD in mini-batch training (as observed by~\citet{kidambi2018insufficiency} and~\citet{shallue2018measuring}). Concurrently, \citet{samuel2019mom} showed that momentum outperforms plain SGD at large batch sizes.

Finally, there are a few works studying average of the iterates, rather than working with the last iterate. This is a classical idea in optimization, where it is known to provide improved convergence~\citep{polyak1992acceleration, bach2013non, dieuleveut2016nonparametric}. However, most of them focused on \emph{tail averaging}, which you have to decide ahead of time the iteration to start accumulating the running averaging. More commonly (especially in deep learning), exponential moving average~\citep{martens2014new} is preferred for its simplicity and ability to handle non-convex landscape. However, no analysis was done especially when mini-batch is used.

\section{Analysis of the Noisy Quadratic Model (NQM)}\label{sec:nqm}
In this section, we work with a \textit{noisy quadratic model} (NQM), a stochastic optimization problem whose dynamics can be simulated analytically, in order to reason about various phenomena encountered in training neural networks. 
In this highly simplified model, we first assume the loss function being optimized is a convex quadratic, with noisy observations of the gradient. 
For analytic tractability, we further assume the noise covariance is codiagonalizable with the Hessian. %
Because we are not interested in modeling overfitting effects, we focus on the online training setting, where the observations are drawn i.i.d.~in every training iteration. 
Under these assumptions, we derive an analytic expression for the risk after any number of steps of SGD with a fixed step size, as well as a dynamic programming method to compute the risk following a given step size schedule.

Convex quadratics may appear an odd model for a complicated nonconvex optimization landscape. However, one obtains a convex quadratic objective by linearizing the network's function around a given weight vector and taking the second-order Taylor approximation to the loss function (assuming it is smooth and convex). Indeed, recent theoretical works~\citep{jacot2018neural, du2018gradient, zhang2019fast} show that for wide enough networks, the weights stay close enough to the initialization for the linearized approximation to remain accurate. Empirically, linearized approximations closely match a variety of training phenomena for large but realistic networks~\citep{lee2019wide}.

\subsection{Problem Setup}\label{sec:prob-setup}
\begin{wrapfigure}[11]{R}{0.35\textwidth}
    \centering
    \vspace{-0.6cm}
    \includegraphics[width=1.9in]{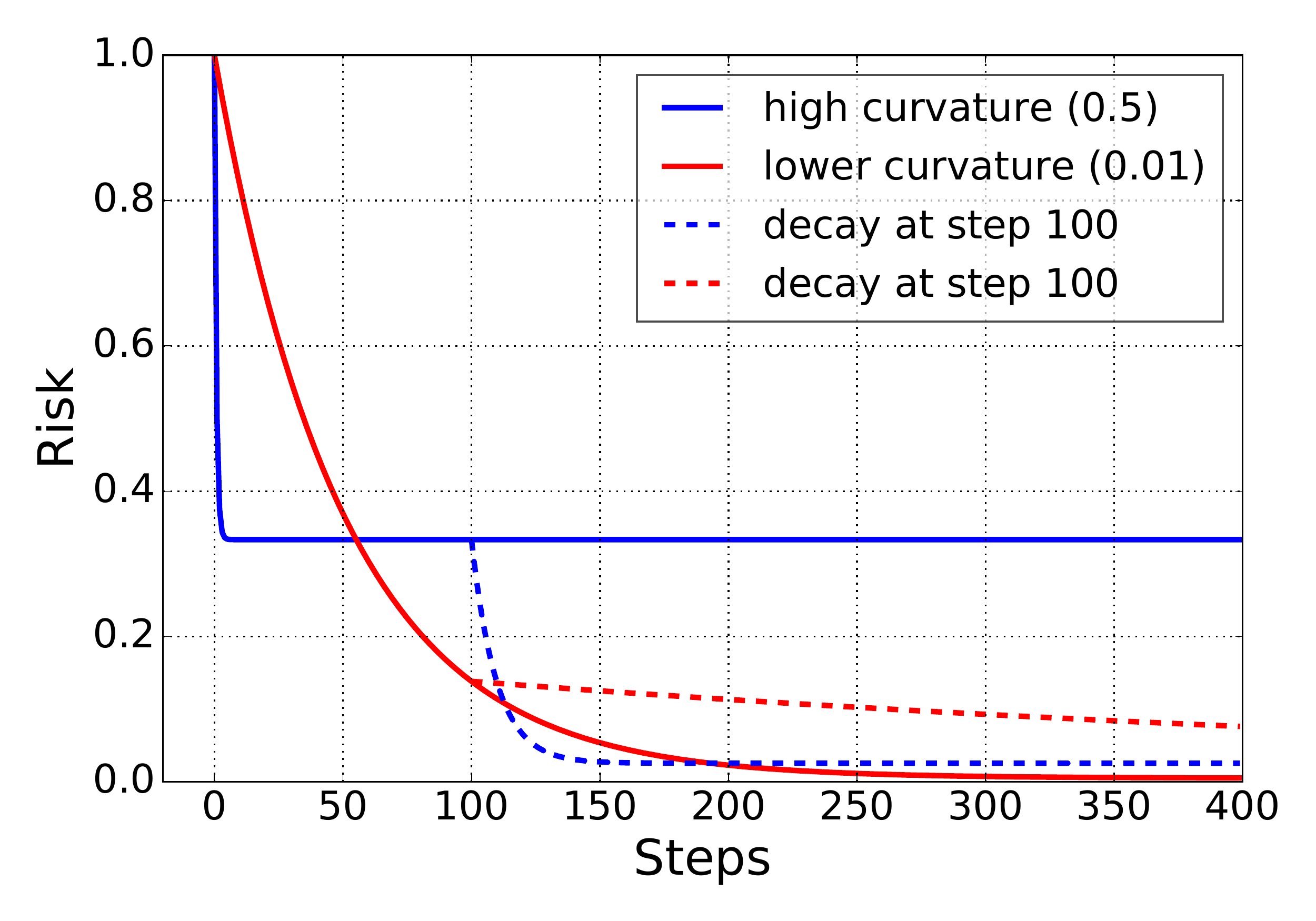}
    \vspace{-0.6cm}
    \caption{Cartoon of the evolution of risk for different coordinates with and without learning rate decay.}
    \label{fig:cartoon}
    \vspace{-0.4cm}
\end{wrapfigure}
We now introduce the noisy quadratic model~\citep{schaul2013no, martens2014new, wu2018understanding}, where the true function being optimized is a convex quadratic.
Because we analyze rotation-invariant and translation-invariant optimizers such as SGD and heavy-ball momentum, we assume without loss of generality that the quadratic form is diagonal, and that the optimum is at the origin. Hence, our exact cost function decomposes as a sum of scalar quadratic functions for each coordinate:
\begin{equation}
    \loss(\params) = \frac{1}{2} \params^\top \hessian \params = \frac{1}{2} \sum_{i=1}^d h_i \theta_i^2 \triangleq \sum_{i=1}^d \ell(\theta_i) .
\end{equation}
Without loss of generality, we assume $h_1 \geq h_2 \geq ... \geq h_d$. 
We consider a single gradient query to have the form $g(\params) = \hessian\params + \bm{\epsilon}$ where $\expect[\bm{\epsilon}] = \mathbf{0}$ and $\cov(\bm{\epsilon}) = \covariance$.
To reduce the variance of gradient estimation, we can average over multiple independent queries, which corresponds to "mini-batch training" in neural network optimization. We denote the averaged gradient as $g_B(\params)$ and the covariance $\cov(g_B(\params)) = \covariance/B$, where $B$ is the number of queries (mini-batch size). 

For analytical tractability, we make the nontrivial assumption that $\mathbf{H}$ and $\mathbf{C}$ are codiagonalizable. (Since $\hessian$ is diagonal, this implies that $\mathbf{C} = \mathrm{diag}(c_1, \ldots, c_d)$.) See Section \ref{sec:hessian-covariance} for justification of this assumption.
Under gradient descent with fixed step size $\alpha$, each dimension evolves independently as
\begin{equation}
    \theta_i(t+1) = (1 - \lr h_i) \theta_i(t) + \lr \sqrt{c_i / B} \epsilon_i ,
\end{equation}
where $\lr$ is the learning rate and $\epsilon_i$ is zero-mean unit variance iid noise. %
By treating $\theta_i$ as a random variable, we immediately obtain the dynamics of its mean and variance.
\begin{equation}\label{eq:exp-var-dynamics}
\begin{aligned}
    \expect\left[\theta_i(t+1) \right] = (1 - \lr h_i)\expect\left[\theta_i(t) \right] , \;
    \vars\left[\theta_i(t+1)\right] = (1 - \lr h_i)^2 \vars\left[\theta_i(t) \right] + \frac{\lr^2 c_i}{B} .
\end{aligned}
\end{equation}
Based on eqn.~\eqref{eq:exp-var-dynamics}, the expected risk after $t$ steps in a given dimension $i$ is
\begin{equation}\label{eq:loss-plain-sgd}
    \expect\left[\ell(\theta_i(t)) \right] = \underbrace{(1 - \lr h_i)^{2t}}_{\text{convergence rate}} \expect\left[\ell(\theta_i(0)) \right]  + \left(1 - (1 - \lr h_i)^{2t}\right)\underbrace{\frac{\lr c_i}{2B (2 - \lr h_i)}}_{\text{steady state risk}} ,
\end{equation}
where we have assumed that $\lr h_i \leq 2$. (Note that this can be seen as a special case of the convergence result derived for convex quadratics in \citet{martens2014new}.)

Remarkably, each dimension converges exponentially to a steady state risk. %
Unfortunately, there is a trade-off in the sense that higher learning rates (up to $1/h_i$) give faster convergence to the steady state risk, but also produce higher values of the steady-state risk. The steady state risk also decreases proportionally to increases in batch size; this is important to note because in the following subsections, we will show that traditional acceleration techniques (e.g., momentum and preconditioning) help improve the convergence rate at the expense of increasing the steady state risk. Therefore, the NQM implies that momentum and preconditioning would benefit more from large-batch training compared to plain SGD, as shown in later sections.

\subsection{Momentum Accelerates Training at Large Batch Sizes}
\label{sec:role-momentum}
Applied to the same noisy quadratic model as before, the update equations for momentum SGD are:
\begin{equation}
\begin{aligned}
    m_i(t+1) &= \beta m_i(t) + h_i \theta_i(t) + \sqrt{c_i/B} \epsilon_i ,\\
    \theta_i(t+1) & = \theta_i(t) - \lr m_i(t+1) .
\end{aligned}
\end{equation}
We show in the following theorem (see Appendix \ref{app:momentum_proof} for proof) that momentum SGD performs similarly to plain SGD in the regime of small batch sizes but helps in the large-batch regime, which can be viewed as a near-deterministic optimization problem. 
\begin{thm}\label{thm:dynamics}
Given a dimension index $i$, and $0 \leq \beta < 1$ with $\beta \neq (1 - \sqrt{\alpha h_i})^2$, the expected risk at time $t$ associated with that dimension satisfies the upper bound
\begin{equation}\label{eq:dynamics}
    \expect\left[\ell(\theta_i(t)) \right] \leq \left( \frac{(r_1^{t+1} - r_2^{t+1}) - \beta (r_1^{t} - r_2^{t}))}{r_1 - r_2}\right)^2\expect\left[\ell(\theta_i(0)) \right]  + \frac{(1 + \beta) \lr c_i}{2B(2\beta + 2 - \lr h_i) (1 - \beta)} ,
\end{equation}
where $r_1$ and $r_2$ (with $r_1 \geq r_2$) are the two roots of the quadratic equation $x^2 - (1 - \lr h_i + \beta)x + \beta = 0$. 
\end{thm}
As with plain SGD (c.f.~eqn.~\eqref{eq:loss-plain-sgd}), the loss associated with each dimension can be expressed as the sum of two terms, where the first one decays exponentially and corresponds to the behavior of the deterministic version of the algorithm, and the second remains constant. 

\begin{wrapfigure}[14]{R}{0.4\textwidth}
	\vspace{-0.5cm}
    \centering
    \includegraphics[width=2.2in]{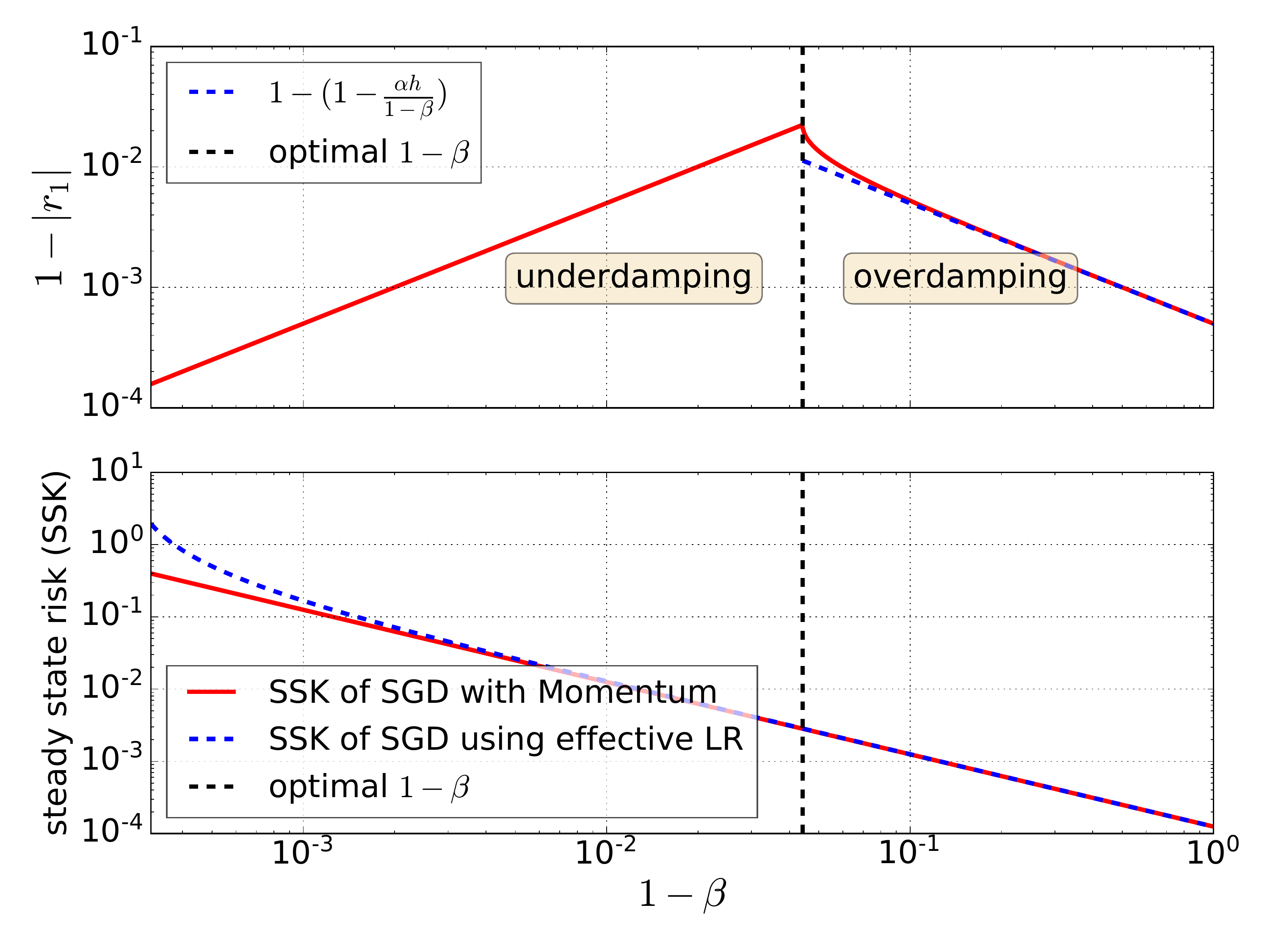}
    \vspace{-0.6cm}
    \caption{Convergence rate and steady state risk (SSK) as a function of momentum for a single dimension with $\alpha h = 0.0005$ and batch size $B = 1$.}
    \label{fig:momentum}
    \vspace{-0.4cm}
\end{wrapfigure}

Following the existing treatment of the deterministic version of the algorithm~\citep{chiang1974fundamental, qian1999momentum, yang2018physical, goh2017momentum}, we divide our analysis two cases: \emph{overdamping} and \emph{underdamping}. In the case of overdamping, where $\beta < (1 - \sqrt{\lr h_i})^2$, both roots $r_1$ and $r_2$ are real and therefore the convergence rate is determined by the larger one (i.e.~$r_1$), which has the value 
\begin{equation}\label{eq:root}
    r_1 = \frac{1 - \lr h_i + \beta + \sqrt{(1 - \beta)^2 - 2(1+\beta)\lr h_i + \alpha^2 h_i^2}}{2}
\end{equation}
With a fixed learning rate, the steady state risk will be constant, and the best achievable expected risk will be lower bounded by it. Thus, to achieve a certain target loss we must either drive the learning rate down, or the batch size up. Assuming a small batch size and a low target risk, we are forced to pick a small learning rate, in which case one can show\footnote{To see this, note that the term in the square root of eqn.~\eqref{eq:root} for $r_1$ can be written as $(1-\beta -  \nicefrac{(1+\beta)\lr h_i}{1-\beta})^2 + \bigO(\alpha^2 h_i^2)$. Dropping the $\bigO(\alpha^2 h_i^2)$ term and simplifying gives the claimed expression for $r_1$.} that $r_1 \approx 1 - \nicefrac{\lr h}{1 - \beta}$. In Figure~\ref{fig:momentum} we plot the convergence rate as a function of $\beta$, and we indeed observe that the convergence rate closely matches $1 - \nicefrac{\alpha h}{1 - \beta}$, assuming a relative small learning rate. We further note that the convergence rate and steady state risk of eqn.~\eqref{eq:dynamics} are the same as the ones in plain SGD (eqn.~\eqref{eq:loss-plain-sgd}), except that they use an "effective learning rate" of $\nicefrac{\lr}{1-\beta}$.
To help validate these predictions, in Appendix~\ref{app:deterministc-term} we provide a comparison of momentum SGD with plain SGD using the effective learning rate. %

In the case of underdamping where $\beta > (1 - \sqrt{\lr h_i})^2$, both $r_1$ and $r_2$ will be complex and have norm $\sqrt{\beta}$. We note that the optimal $\beta$ should be equal to or smaller than $(1 - \sqrt{\lr h_d})^2$, since otherwise all dimensions are under-damped, and we can easily improve the convergence rate and steady state risk by reducing $\beta$. 

Next we observe that the convergence of the total loss will \emph{eventually} be dominated by the slowest converging dimension (which corresponds to the smallest curvature $h_d$), and this will be in the overdamping regime as argued above. By our analysis of the overdamping case, we can achieve the same convergence rate for this dimension by simply replacing the learning rate $\lr$ in the bound for plain SGD (eqn.~\eqref{eq:loss-plain-sgd}) with the effective learning rate $\nicefrac{\lr}{1 - \beta}$.

So while momentum gives no long-term training acceleration for very low fixed learning rates (which we are forced to use when the batch size is small), we note that it can help in large-batch training. 
With $\beta > 0$, the steady state risk roughly amplifies by a factor of $\nicefrac{1}{1 - \beta}$, and we note that steady state risk also decreases proportionally to increases in batch size.
Therefore, we expect momentum SGD to exhibit perfect scaling up to larger batch sizes than plain SGD.

\subsection{Preconditioning Further Extends Perfect Scaling to Larger Batch Sizes}
Many optimizers, such as Adam and K-FAC, can be viewed as preconditioned gradient descent methods. In each update, the gradient is rescaled by a PSD matrix $\precon^{-1}$, called the preconditioner. 
\begin{equation}\label{eq:precon-update-eq}
    \params(t+1) = \params(t) - \lr \precon^{-1} \left[\hessian \params + \bm{\epsilon} \right] .
\end{equation}
In lieu of trying to construct noisy quadratic analogues of particular optimizers, we analyze preconditioners of the form $\precon = \hessian^p$ with $0 \leq p \leq 1$. Note that $\precon$ remains fixed throughout training since the Hessian $\hessian$ is constant in the NQM. We can recover standard SGD by setting $p = 0$. 

Conveniently, for our NQM, the dynamics of preconditioned SGD are equivalent to the SGD dynamics in an NQM with Hessian $\transHessian = \precon^{-1/2} \hessian \precon^{-1/2}$ and gradient covariance $\transCovariance = \precon^{-1/2} \covariance \precon^{-1/2}$. Hence, the dynamics can be simulated using eqn.~(\ref{eq:loss-plain-sgd}), exactly like the non-preconditioned case.
We immediately obtain the following bound on the risk: %
\begin{equation}\label{eq:precon-total-loss}
    \expect\left[\loss(\params(t)) \right] \leq \sum_{i=1}^d (1 - \lr h_i^{(1-p)})^{2t} \expect\left[\ell(\theta_i(0)) \right]  + \sum_{i=1}^d \frac{\lr c_i h_i^{-p}}{2B (2 - \lr h_i^{1-p} )} . %
\end{equation}
To qualitatively understand the effect of preconditioning, first consider the first term in eqn.~\eqref{eq:precon-update-eq}. The convergence of this term resembles that of gradient descent on a deterministic quadratic, which (with optimal $\alpha\approx 2/\tilde{h}_1$) converges exponentially at a rate of approximately $2/\tilde{\kappa}$, where $\tilde{\kappa}=\tilde{h}_1/\tilde{h}_d$ is the condition number of the transformed problem. Since $\tilde{\kappa} = \kappa^{1-p}$, this implies a factor of $\kappa^p$ improvement in the rate of convergence. Hence, for near-deterministic objectives where the first term dominates, values of $p$ closer to 1 correspond to better preconditioners, and result in much faster convergence.
Unfortunately, there is no free lunch, as larger values of $p$ will also increase the second term (steady state risk). Assuming an ill-conditioned loss surface ($\kappa \gg 1$), the steady state risk of each dimension becomes
\begin{equation}
    \frac{1}{2B}\frac{\lr c_i h_i^{-p}}{2 - \lr h_i^{(1-p)}} \approx \frac{c_i}{2B h_1} \frac{(h_i/h_1)^{-p}}{1 - (h_i/h_1)^{(1-p)}} ,
\end{equation}
which is a monotonically increasing function with respect to $p$. %
Even without this amplification effect, the steady state risk will eventually become the limiting factor in the minimization of the expected risk. One way to reduce the steady state risk, apart from using Polyak averaging~\citep{polyak1992acceleration} or decreasing the learning rate (which will harm the rate of convergence), is to increase the batch size. This suggests that the benefits of using stronger preconditioners will be more clearly observed for larger batch sizes, which is an an effect that we empirically demonstrate in later sections.

\subsection{Exponential Moving Average Reduces Steady State Risk}
Following the same procedure as previous two sections, we analyze exponential moving averages (EMA) on our NQM. The update rule of EMA can be written as
\begin{equation}
\begin{aligned}
    \params(t+1) &= \params(t) - \lr  \left[\hessian \params + \bm{\epsilon} \right], \\
    \tilde{\params}(t+1) &=  \gamma \tilde{\params}(t) + (1 - \gamma) \params(t+1).
\end{aligned}
\end{equation}
The averaged iterate $\tilde{\params}$ is used at test time. The computational overhead is minimal (storing an additional copy of the parameters, plus some cheap arithmetic operations). We now show that EMA outperforms plain SGD by reducing the steady state risk term.
\begin{thm}\label{thm:ema-dynamics} 
Given a dimension index $i$, and $0 \leq \gamma < 1$, the expected risk at time t associated with that dimension satisfies the upper bound
\begin{equation}\label{eq:ema-dynamics}
\begin{aligned}
    \expect\left[\ell(\tilde{\theta}_i(t)) \right] &\leq \left( \frac{(r_1^{t+1} - r_2^{t+1}) - \gamma(1-\alpha h_i) (r_1^{t} - r_2^{t}))}{r_1 - r_2}\right)^2\expect\left[\ell(\theta_i(0)) \right] \\ &+ \frac{\alpha c_i}{2B (2 - \alpha h_i)} {\color{red}\frac{(1-\gamma)(1+ (1 - \alpha h_i) \gamma)}{(1+\gamma)(1 - (1 - \alpha h_i) \gamma)}} ,
\end{aligned}
\end{equation}
where $r_1 = 1 - \lr h_i$ and $r_2 = \gamma$.
\end{thm}
By properly choosing an averaging coefficient $\gamma < 1 - \alpha h_d$ such that $r_1 > r_2$, one can show that EMA reduces the steady state risk without scarificing the convergence rate. To see this, we note that the {\color{red}red} part of eqn.~\eqref{eq:ema-dynamics} is strictly less than $1$ given the fact $1 - \lr h_i < 1$ while the other part is exactly the same as the steady state risk of plain SGD.

\subsection{Choice of $\hessian$ and $\covariance$}\label{sec:hessian-covariance}
We've found that the qualitative behavior of optimizers in our NQM depends on the choices of $\hessian$ and $\covariance$. Therefore, we choose matrices motivated by theoretical and empirical considerations about neural net training. First, we set the diagonal entries of $\hessian$ to be $\{\frac{1}{i}\}_{i=1}^d$ for some integer $d$, giving a condition number of $d$. This closely matches the estimated eigenspectrum of the Hessian of a convolutional network (see Figure~\ref{fig:eigenspectra} and Appendix~\ref{app:verification-eigen}), and is also consistent with recent work finding heavy tailed eigenspectra of neural network Hessians~\citep{ubaru2017fast, ghorbani2019investigation}. We choose $d=10^4$, which approximately matches the condition number of the K-FAC Hessian approximation for ResNet8. (Qualitative behaviors were consistent for a wide range of $d$.)

We also set $\covariance = \hessian$ (a nontrivial assumption). %
This was motivated by theoretical arguments that, under the assumption that the implicit conditional distribution over the network's output %
is close to the conditional distribution of targets from the training distribution, the Hessian closely matches the gradient covariance in neural network training~\citep{martens2014new}. Empirically, this relationship appears to hold tightly for a convolutional network and moderately well for a transformer (see Appendix~\ref{app:grad-var}).

\subsection{Information Theoretic Lower Bound}
Since our NQM assumes the infinite data (online optimization) setting, it's instructive to compare the performance of optimizers against an information theoretic lower bound. Specifically, under the assumption that $\hessian = \covariance$, the NQM is equivalent to maximum likelihood estimation of the mean vector for a multivariate Gaussian distribution with covariance $\hessian^{-1}$. Hence, the risk obtained by any optimizer can be bounded below by the risk of the maximum likelihood estimator for the Gaussian, which is $d/2N$, where $d$ is the dimension and $N$ is the total number of training examples visited. We indicate this bound with a dashed black line in our plots.

\subsection{Noisy Quadratic Experiments}
\label{sec:nqm-exp}
In this section, we simulate noisy quadratic optimization using the closed-form dynamics. Our aim is to formulate hypotheses for how different optimizers would behave for neural network optimization. Our main metric is the number of steps required to achieve a target risk. For efficiency, rather than explicitly representing all the eigenvalues of $\hessian$, we quantize them into 100 bins and count the number of eigenvalues in each bin. Unless otherwise specified, we initialize $\params$ as $\normal(\bm{0}, \iden)$ and use a target risk of 0.01. (The results don't seem to be sensitive to either the initial variance or the target risk; some results with varying target risk thresholds are shown in Appendix~\ref{app:loss-thres}).

\begin{figure}[t]
\centering
\begin{subfigure}[t]{0.32\textwidth}
    \centering
    \includegraphics[height=1.15in]{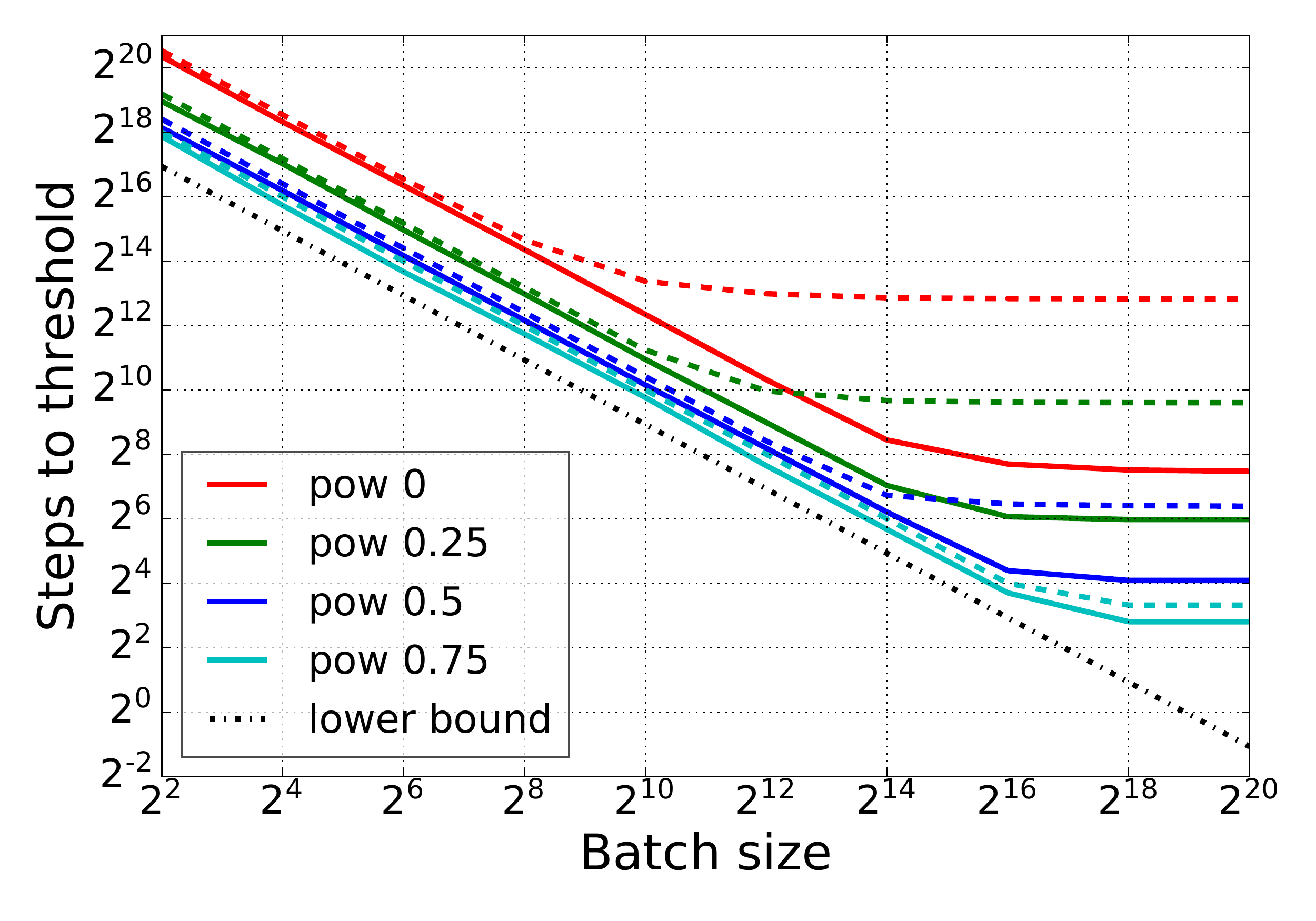}
    \vspace{-0.2cm}
    \caption{\small Momentum and Preconditioning}
    \label{fig:sgd_momentum}
\end{subfigure}
\begin{subfigure}[t]{0.32\textwidth}
    \centering
    \includegraphics[height=1.15in]{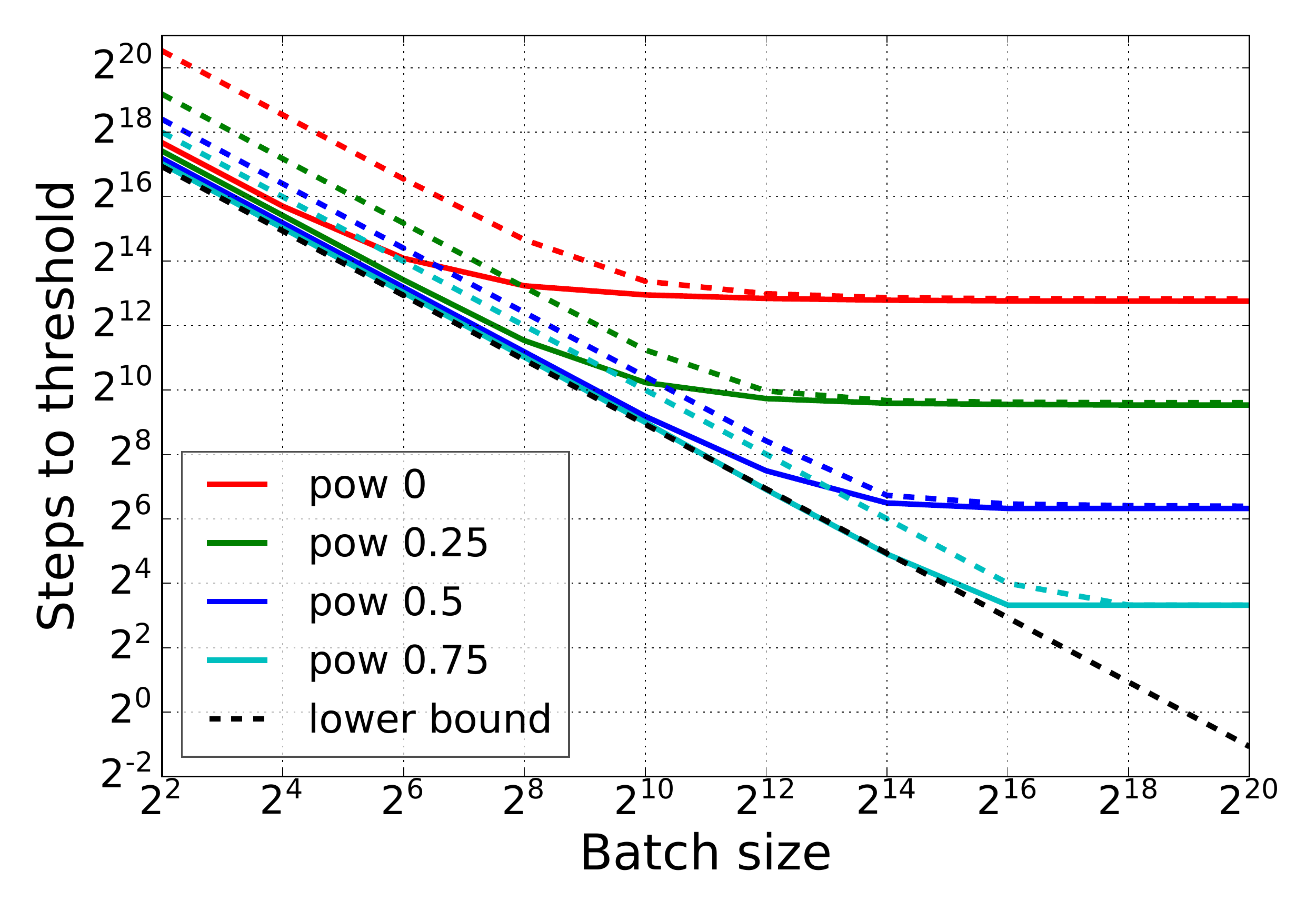}
    \vspace{-0.2cm}
    \caption{\small Fixed LR vs.~Schedules}
    \label{subfig:pwc}
\end{subfigure}
\begin{subfigure}[t]{0.32\textwidth}
    \centering
    \includegraphics[height=1.15in]{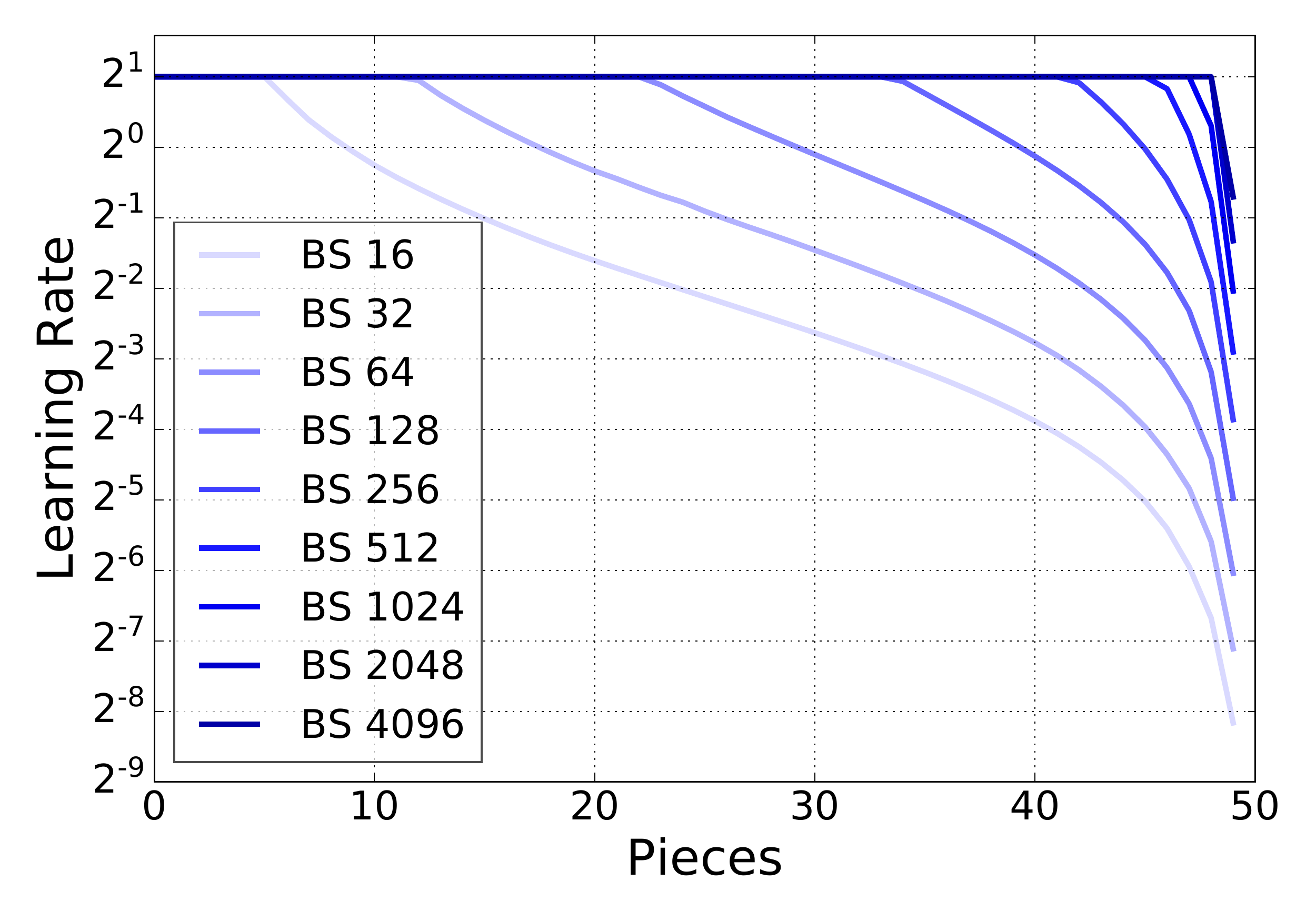}
    \vspace{-0.2cm}
    \caption{Optimized LR Schedules}
    \label{subfig:pwc_optlr}
\end{subfigure}
\caption{\textbf{(a) Effects of momentum and preconditioning.} Steps required to reach target loss as a function of batch size under different preconditioning power. Solid lines are momentum SGD while dashed lines are plain SGD. The black dashed line is the information theoretic lower bound. \textbf{(b) Effect of learning rate decay.} The solid lines use the optimized piecewise constant scheme, which are shown in \textbf{(c)} for power $0$. The dashed curves in \textbf{(b)} are plain SGD for comparison. We observe that learning rate schedules close most of the gap between the fixed learning rate performance and the information theoretic lower bound.}
\vspace{-0.4cm}
\label{fig:optimal_lr} 
\end{figure}
\subsubsection{Effect of Momentum, Preconditioning and Exponential Moving Average}
\label{sec:effect-momentum-precon}
We first experiment with momentum and varying preconditioner powers on our NQM. We treat both the (fixed) learning rate $\alpha$ and momentum decay parameter $\beta$ as hyperparameters, which we tune using a fine-grained grid search. %

Consistent with the empirical results of~\citet{shallue2018measuring}, each optimizer shows two distinct regimes: a small-batch (stochastic) regime with perfect linear scaling, and a large-batch (deterministic) regime insensitive to batch size. We call the phase transition between these regimes the \emph{critical batch size}. Consistent with the analysis of Section~\ref{sec:role-momentum} and the observations of~\citet{smith2017don, shallue2018measuring, kidambi2018insufficiency}, the performance of momentum-based optimizers matches that of the plain SGD methods in the small-batch regime, but momentum increases the critical batch size and gives substantial speedups in the large batch regime.
Preconditioning also increases the critical batch size and gives substantial speedups in the large batch regime, but interestingly, also improves performance by a small constant factor even for very small batches. Combining momentum with preconditioning extends both of these trends. %

\begin{wrapfigure}[10]{R}{0.33\textwidth}
	\vspace{-0.7cm}
    \centering
    \includegraphics[height=1.2in]{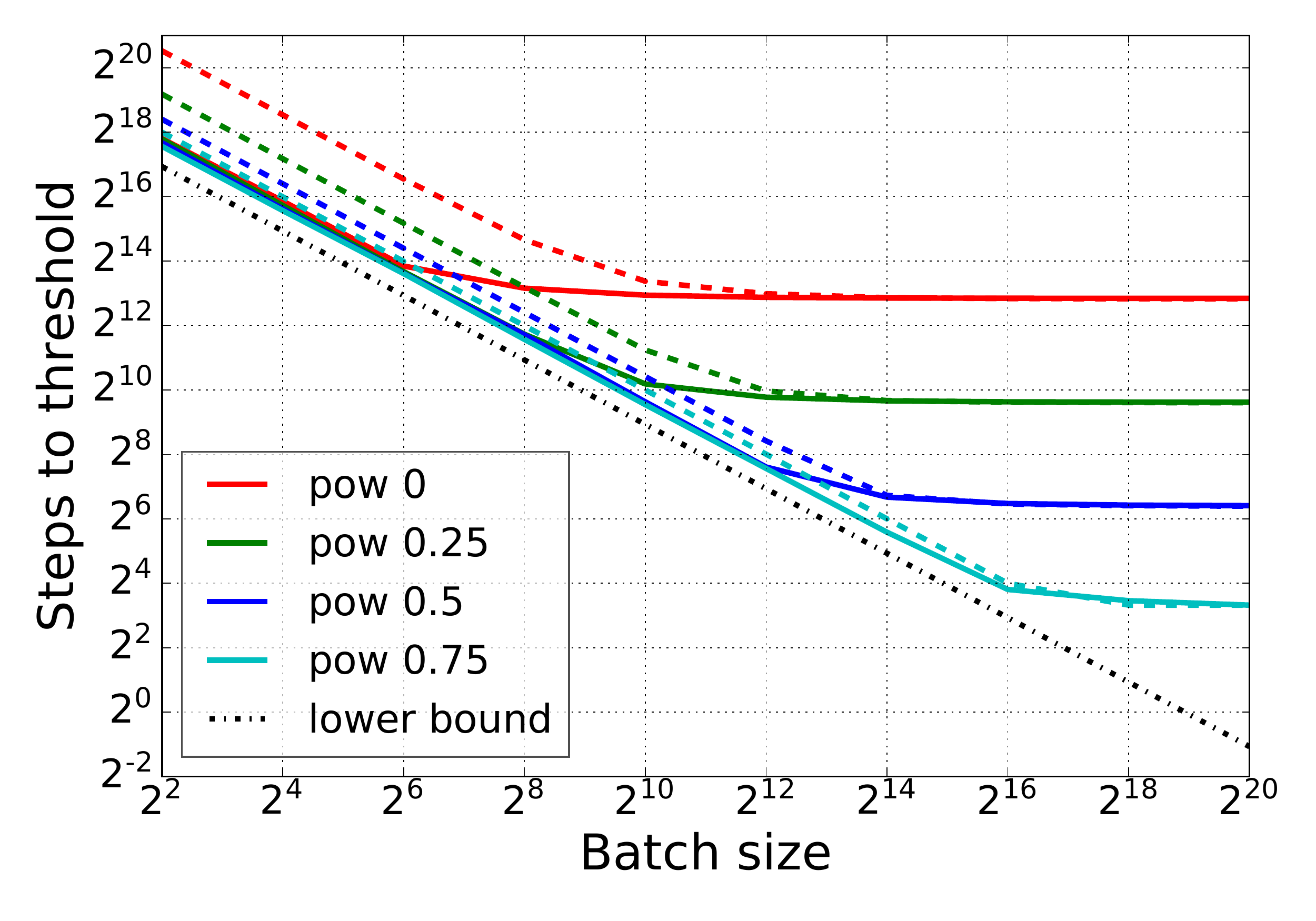}
    \vspace{-0.35cm}
    \caption{Effects of exponential moving average (EMA). Solid lines are SGD with EMA while dashed lines are plain SGD.}
    \label{fig:training-target}
\end{wrapfigure}
We next experiment with EMA and varying preconditioning powers on our NQM. Following the same procedure as before, we tune both learning rate $\alpha$ and averaging coefficient $\gamma$ using grid search. As expected, EMA reduces the number of steps required especially for plain SGD with preconditioning power $0$. Another interesting observation is that EMA becomes redundant in the large batch (near-deterministic) regime since the main effect of EMA is reducing the steady-state risk, which can also be done by increasing the batch size. This implies that EMA would reduce the critical batch size and therefore achieve the same amount of acceleration with less computation.

\subsubsection{Optimal Learning Rate and Decay Scheme }
\label{sec:lrate-scheme}
In the NQM, we can calculate the optimal constant learning rate given a specific batch size. %
Figure~\ref{fig:optimal_lr} shows the optimal learning rate as a function of batch size for a target risk of $0.01$.
Notably, the optimal learning rate of plain (preconditioned) SGD (Figure~\ref{subfig:plain-sgd-optlr}) scales linearly with batch size before it hits the critical batch size, matching the scheme used in~\citet{goyal2017accurate}. The linear scaling also holds for the effective learning rate of momentum SGD. In the small batch regime, the optimal effective learning rate for momentum SGD matches the optimal plain SGD learning rate, suggesting that the momentum and learning rate are interchangeable in the small batch regime.

While a fixed learning rate often works well for simple problems, good performance on the ImageNet benchmark~\citep{russakovsky2015imagenet} requires a carefully tuned schedule. Here we explicitly optimize a piecewise constant learning rate schedule for SGD (with 50 pieces), in terms of the number of steps to reach the loss threshold.\footnote{For a given schedule and number of time steps, we obtain the exact risk using dynamic programming with eqn.~\eqref{eq:exp-var-dynamics}. For stability, the learning rates are constrained to be at most $2/h_1$. For a fixed number of time steps, we minimize this risk using BFGS. We determine the optimal number of time steps using binary search.} 
In Figure~\ref{subfig:pwc}, we show that optimized learning rate schedules help significantly in the small batch regime, consistent with the analysis in~\citet{wu2018understanding}. We observe the same linear scaling as with fixed-learning-rate SGD, but with a better constant factor. In fact, optimized schedules nearly achieve the information theoretic optimum. However, learning rate schedules do not improve at all over fixed learning rates in the large batch regime. Figure~\ref{subfig:pwc_optlr} shows optimized schedules for different batch sizes; interestingly, they maintain a large learning rate throughout training followed by a roughly exponential decay, consistent with commonly used neural network training schedules. 
Additionally, even though the different batch sizes start with the same learning rate, their final learning rates at the end of training scale linearly with batch size (see Figure~\ref{fig:pwc_final_lrate} in Appendix~\ref{app:final-lrate}).

\begin{table}[t]
\begin{center}
\begin{small}
\resizebox{0.9\textwidth}{!}{
\begin{tabular}{p{1.2cm}p{1.0cm}p{3.0cm}p{5.0cm}p{1.8cm}}
\toprule
\textbf{Data Set} & \textbf{Size} & \textbf{Model} & \textbf{Remarks} & \textbf{LR}   \\ 
\midrule	
MNIST & 55,000 & \multirow{2}{*}{Simple CNN} & \multirow{2}{*}{\parbox{4.5cm}{Same as~\citet{shallue2018measuring} except without dropout regularization.}} & \multirow{2}{*}{Constant} \\
\cmidrule{1-2}
FMNIST & 55,000 & & & \\
\midrule
\multirow{3}{*}{CIFAR10} & \multirow{3}{*}{45,000} & ResNet8 without BN & Same as~\cite{shallue2018measuring}. & Constant \\               
\cmidrule{3-5}
 & & ResNet32 with BN & Ghost batch norm is used. & Linear Decay \\           
\cmidrule{3-5}
 & & VGG11 with BN & Ghost batch norm is used. & Linear Decay \\           
\midrule
LM1B & $\sim$30M & {Two-layer Transformer} & Shallow model in~\citet{shallue2018measuring} & Constant \\          
\bottomrule
\end{tabular}}

\end{small}
\end{center}
\caption{\textbf{Data sets and models used in our experiments.} See Appendix~\ref{app:model-details} for full details.}
\vspace{-0.6cm}
\label{tab:datasets_and_models}
\end{table}

\begin{figure}[t]
\centering
\begin{subfigure}[t]{0.32\textwidth}
    \centering
    \includegraphics[height=1.18in]{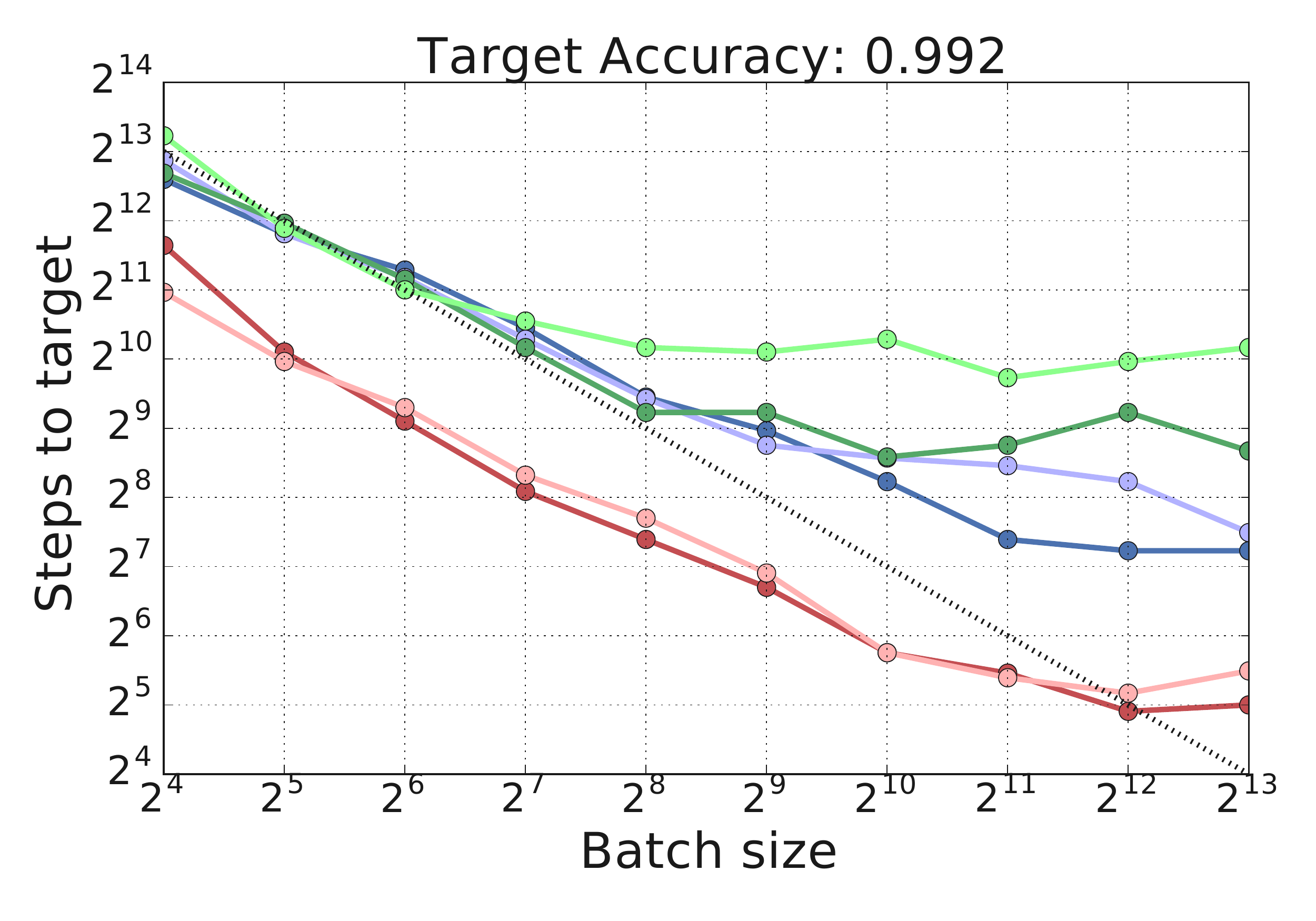}
    \vspace{-0.3cm}
    \caption{{\footnotesize Simple CNN on MNIST}}
    \label{subfig:cnn-mnist}
\end{subfigure}
\begin{subfigure}[t]{0.32\textwidth}
    \centering
    \includegraphics[height=1.18in]{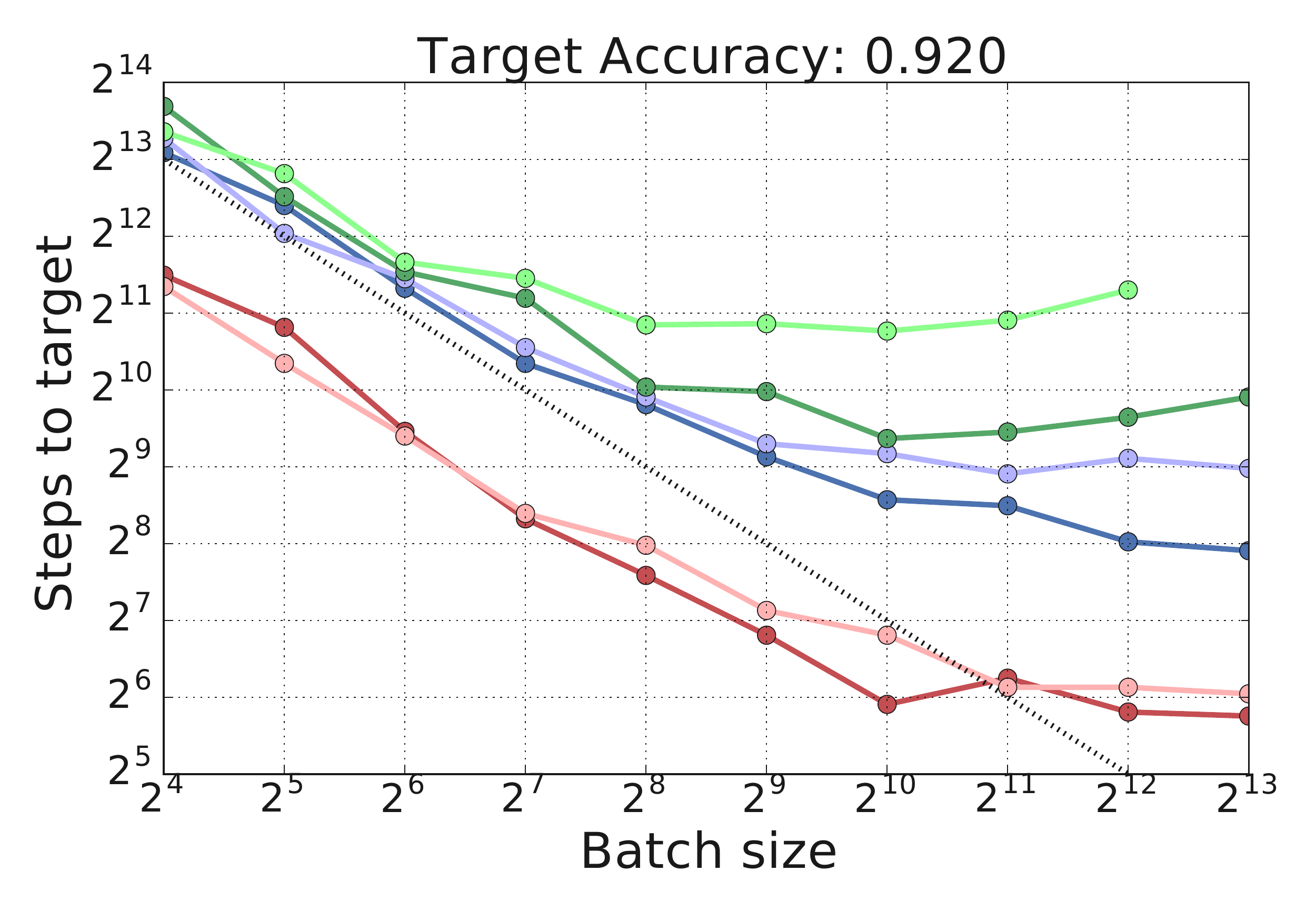}
    \vspace{-0.3cm}
    \caption{{\footnotesize Simple CNN on Fashion MNIST}}
    \label{subfig:cnn-fmnist}
\end{subfigure}
\begin{subfigure}[t]{0.32\textwidth}
    \centering
    \includegraphics[height=1.18in]{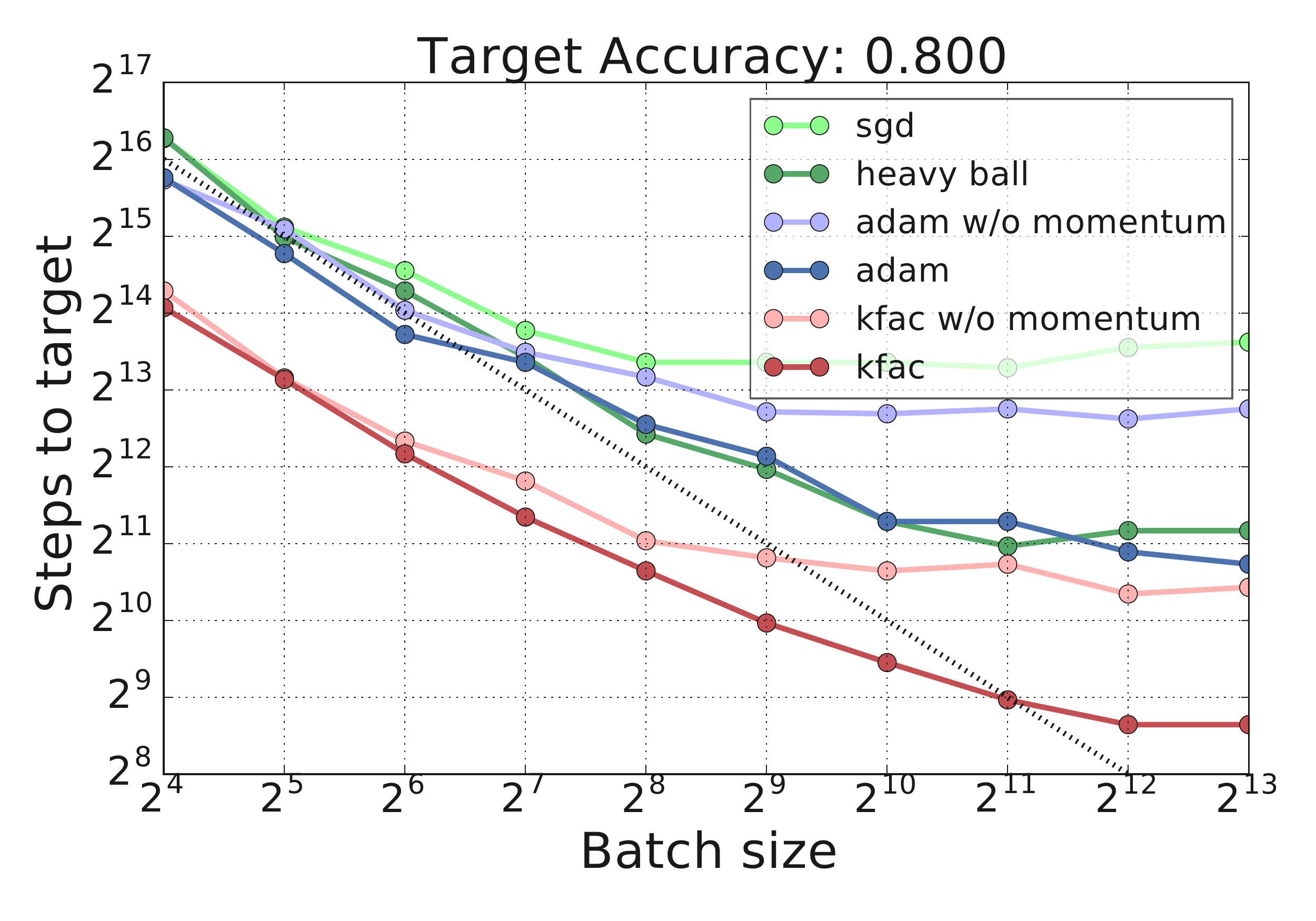}
    \vspace{-0.3cm}
    \caption{{\footnotesize ResNet8 on CIFAR10}}
    \label{subfig:resnet8-cifar}
\end{subfigure}
\begin{subfigure}[t]{0.32\textwidth}
    \centering
    \includegraphics[height=1.18in]{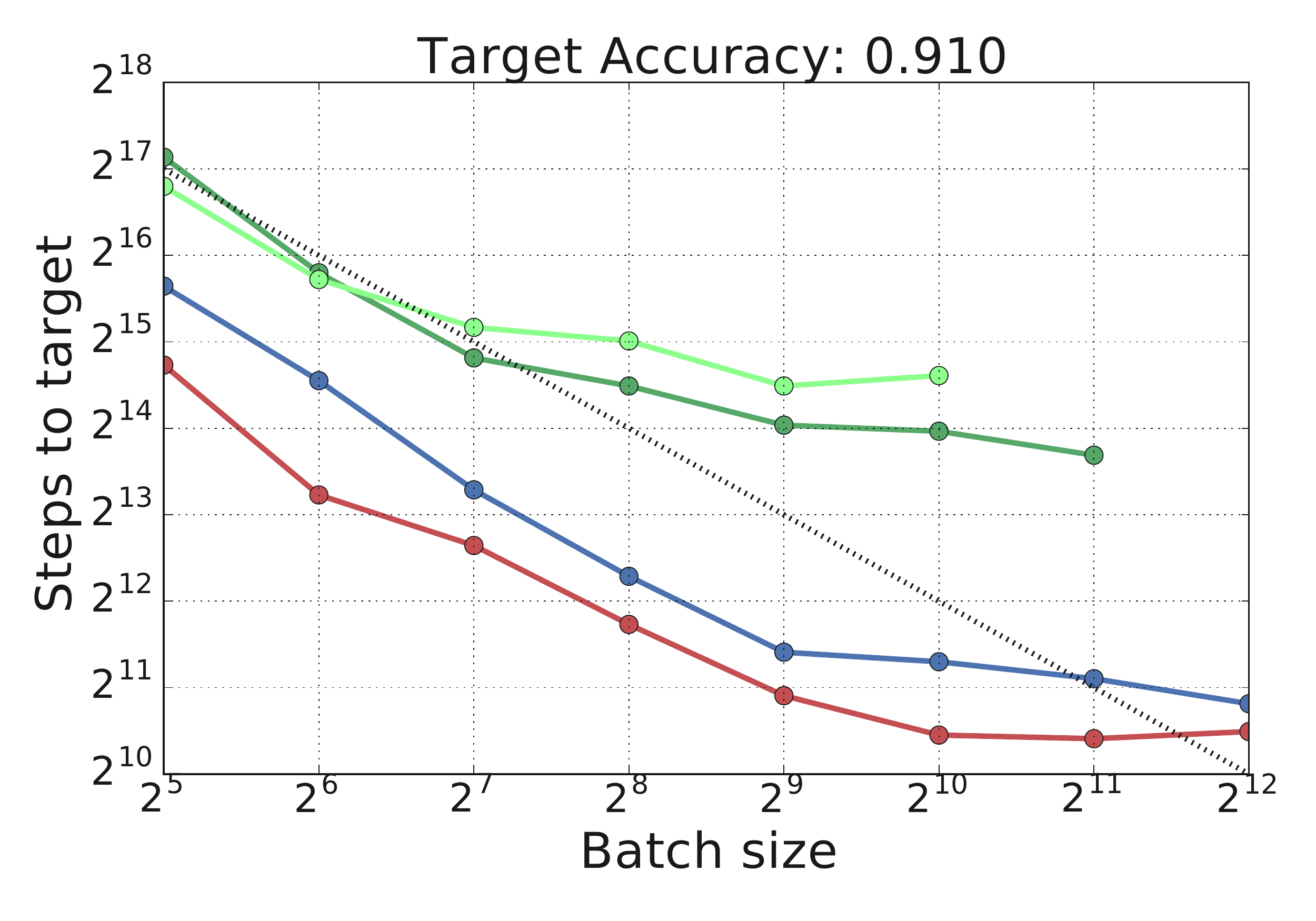}
    \vspace{-0.3cm}
    \caption{{\footnotesize VGG11 on CIFAR10}}
\end{subfigure}
\begin{subfigure}[t]{0.32\textwidth}
    \centering
    \includegraphics[height=1.18in]{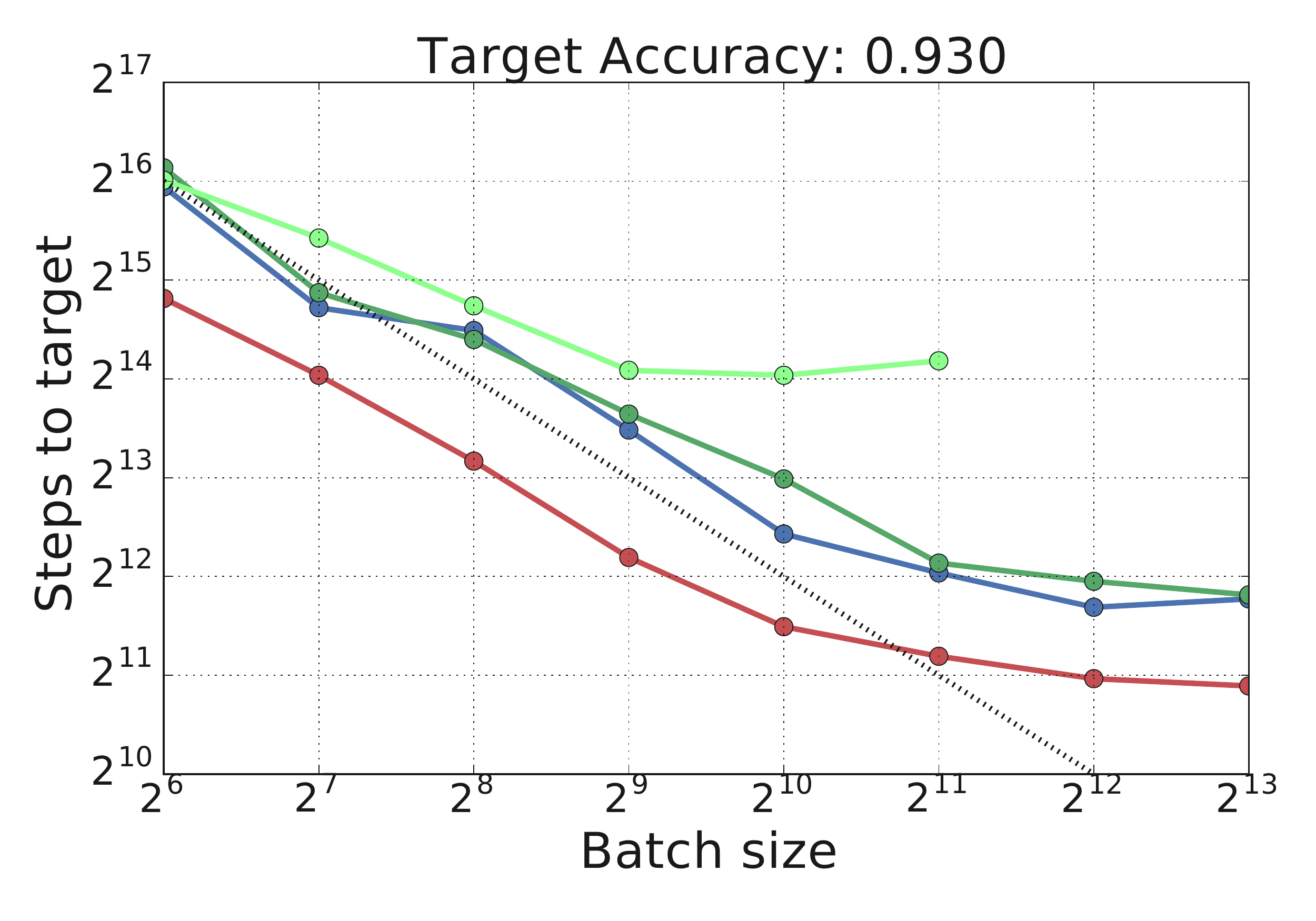}
    \vspace{-0.3cm}
    \caption{{\footnotesize ResNet32 on CIFAR10}}
    \label{subfig:resnet32-cifar}
\end{subfigure}
\begin{subfigure}[t]{0.32\textwidth}
    \centering
    \includegraphics[height=1.18in]{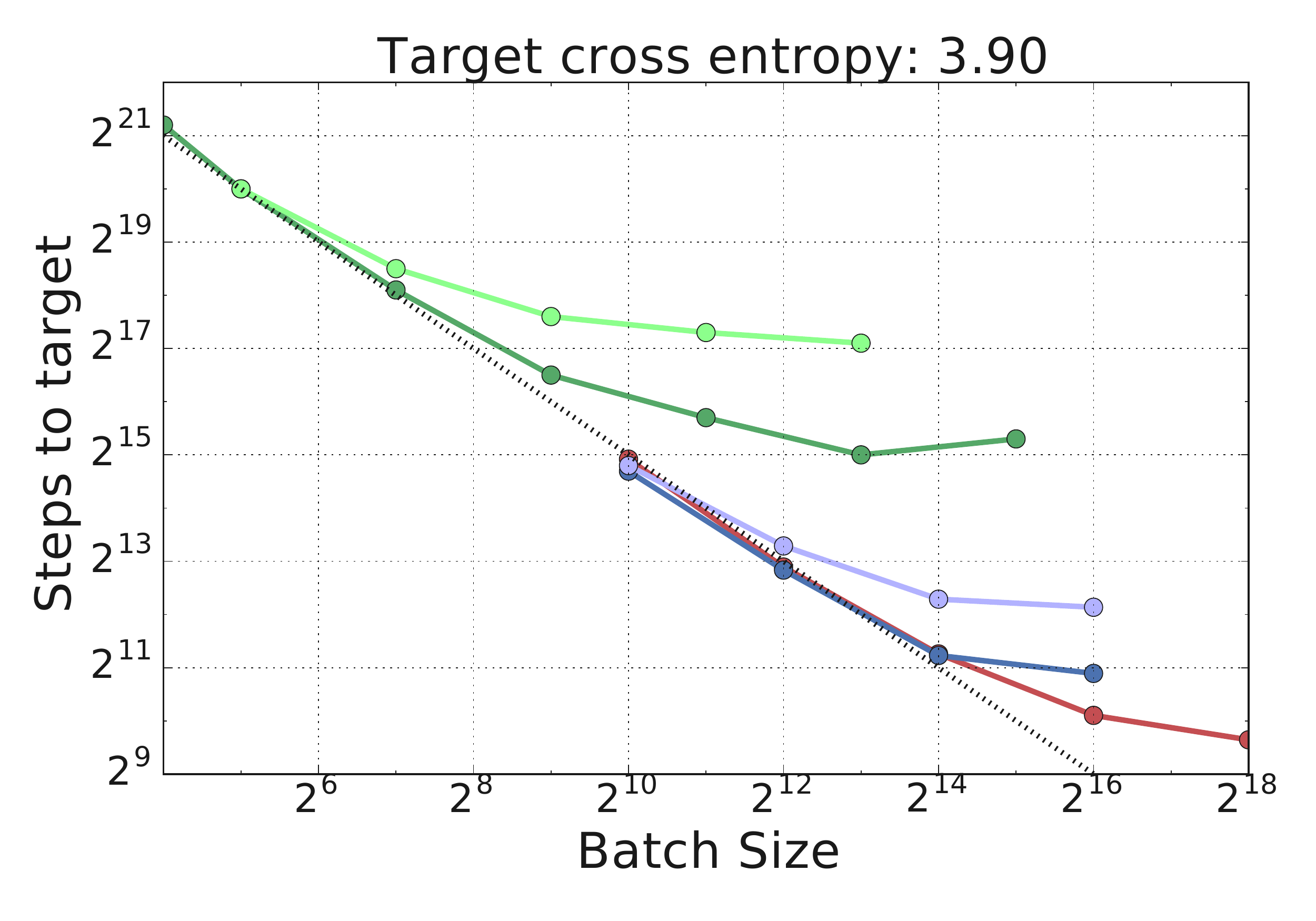}
    \vspace{-0.3cm}
    \caption{{\footnotesize Transformer on LM1B}}
    \label{subfig:transformer-lm1b}
\end{subfigure}
\caption{\textbf{Empirical relationship between batch size and steps to result.} Key observations: 1) momentum SGD has no benefit over plain SGD at small batch sizes, but extends the perfect scaling to larger batch sizes; 2) preconditioning also extends perfect scaling to larger batch sizes, i.e. K-FAC > Adam > momentum SGD. This is most noticeable in the Transformer model; 3) preconditioning (particularly K-FAC) reduces the number of steps needed to reach the target even for small batch sizes. All of these agree with the predictions by NQM.}
\vspace{-0.4cm}
\label{fig:large-scale} 
\end{figure}
\section{Neural Network Experiments}

We investigated whether the predictions made by the NQM hold in practice by running experiments with five neural network architectures across three image classification tasks and one language modeling task (see Table~\ref{tab:datasets_and_models}). For each model and task, we compared a range of optimizers: SGD, momentum SGD, Adam (with and without momentum), and K-FAC (with and without momentum). For K-FAC, preconditioning is applied before momentum.
See Appendix \ref{app:experiments} for more details.

The primary quantity we measured is the number of steps required to reach a target accuracy (for image classification tasks) or cross entropy (for language modeling). Unless otherwise specified, we measured steps to target on the validation set. We chose the target metric values based on an initial set of experiments with practical computational budgets. For each model, task, optimizer, and batch size, we independently tuned the learning rate $\lr$, the parameters governing the learning rate schedule (where applicable), and optimizer-specific metaparameters (see Appendix \ref{app:optimizer-parameters}). 
We manually chose the search spaces based on our initial experiments, and we verified after each experiment that the optimal metaparameter values were far from the search space boundaries. We used quasi-random search \citep{BousquetEtAl_LDS_2017} to tune the metaparameters with fixed budgets of non-divergent\footnote{We discarded trials with a divergent training loss, which occurred when the learning rate was too high.} trials (100 for Simple CNN, ResNet8, and Transformer, and 200 for ResNet32 and VGG11).  We chose the trial that reached the target metric value using the fewest number of steps.

\subsection{Critical Batch Size Depends on the Optimizer}\label{sec:validation-target}
Figure~\ref{fig:large-scale} shows the relationship between batch size and steps to target for each model, task, and optimizer. In each case, as the batch size grows, there is an initial period of perfect scaling where doubling the batch size halves the steps to target, but once the batch size exceeds a problem-dependent critical batch size, there are rapidly diminishing returns, matching the results of~\citep{goyal2017accurate, mccandlish2018empirical, shallue2018measuring}.
K-FAC has the largest critical batch size in all cases, highlighting the usefulness of preconditioning. Momentum SGD extends perfect scaling to larger batch sizes than plain SGD, but for batch sizes smaller than the plain SGD critical batch size, momentum SGD requires as many steps as plain SGD to reach the target. This is consistent with both the empirical results of~\citet{shallue2018measuring} and our NQM simulations. By contrast, Adam and K-FAC can reduce the number of steps needed to reach the target compared to plain SGD even for the smallest batch sizes, although neither optimizer does so in all cases. Finally, we see some evidence that the benefit of momentum diminishes with preconditioning (Figures~\ref{subfig:cnn-mnist} and~\ref{subfig:cnn-fmnist}), as predicted by our NQM simulations, although we do not see this in all cases (e.g. Figure~\ref{subfig:resnet8-cifar} and~\ref{subfig:transformer-lm1b}).

\subsection{Exponential Moving Average Improves Convergence with Minimal Computation Cost}
\begin{wrapfigure}[10]{R}{0.6\textwidth}
	\vspace{-0.4cm}
    \centering
    \begin{subfigure}[t]{0.295\textwidth}
        \centering
        \includegraphics[height=1.1in]{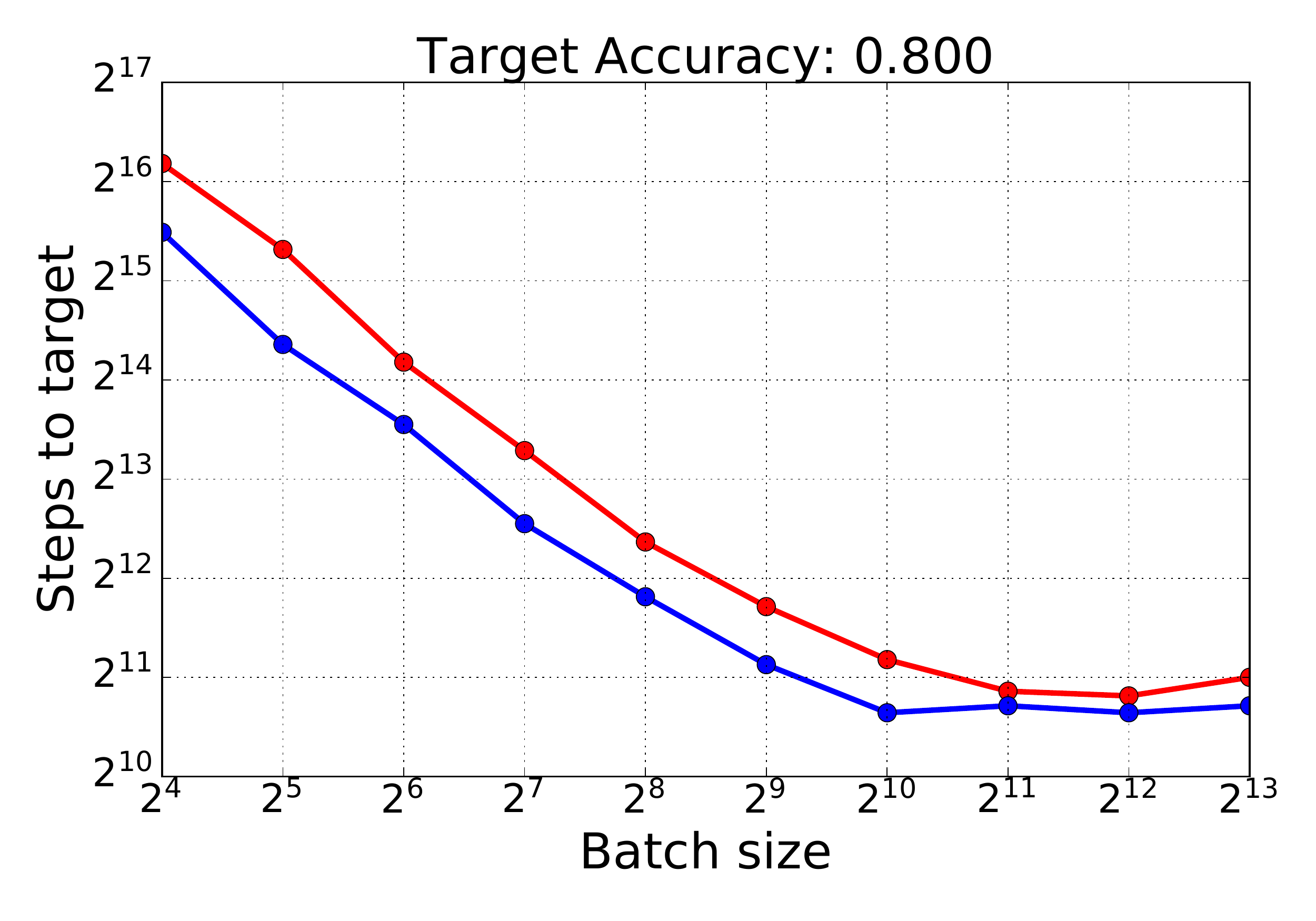}
    \end{subfigure}
    \begin{subfigure}[t]{0.295\textwidth}
        \centering
        \includegraphics[height=1.1in]{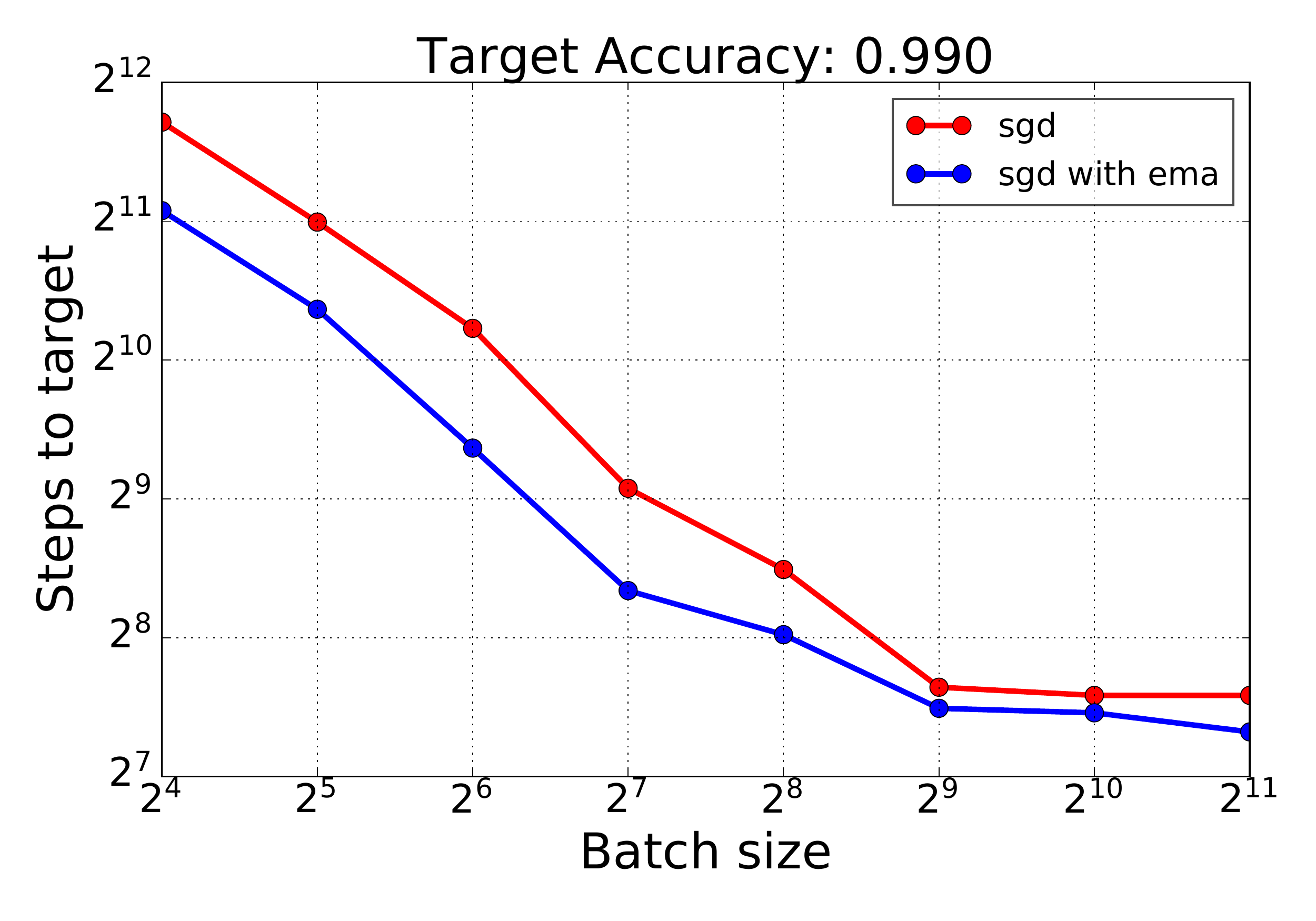}
    \end{subfigure}
    \vspace{-0.6cm}
    \caption{Steps to training accuracy versus batch size. \textbf{Left:} ResNet8 on CIFAR10; \textbf{Right}: Simple CNN on MNIST.}
    \label{fig:ema-real}
    \vspace{-0.4cm}
\end{wrapfigure}
To verify the predictions of NQM on exponential moving average (EMA), we conducted some experiments on comparing EMA with plain SGD. We follow the same protocol of Figure~\ref{fig:large-scale} and report the results in Figure~\ref{fig:ema-real}. As expected, the results on real neural networks closely match our predictions based on NQM analysis. In particular, SGD with EMA appears to reach the same target with fewer steps than plain SGD at small batch sizes, though the benefit of EMA diminishes with large batch sizes. Besides, we note that EMA leads to smaller critical batch sizes and achieves the same acceleration with less computation.

\subsection{Optimal Learning Rate}
\begin{wrapfigure}[9]{R}{0.6\textwidth}
 	\vspace{-0.6cm}
    \centering
    \begin{subfigure}[t]{0.295\textwidth}
        \centering
        \includegraphics[height=1.1in]{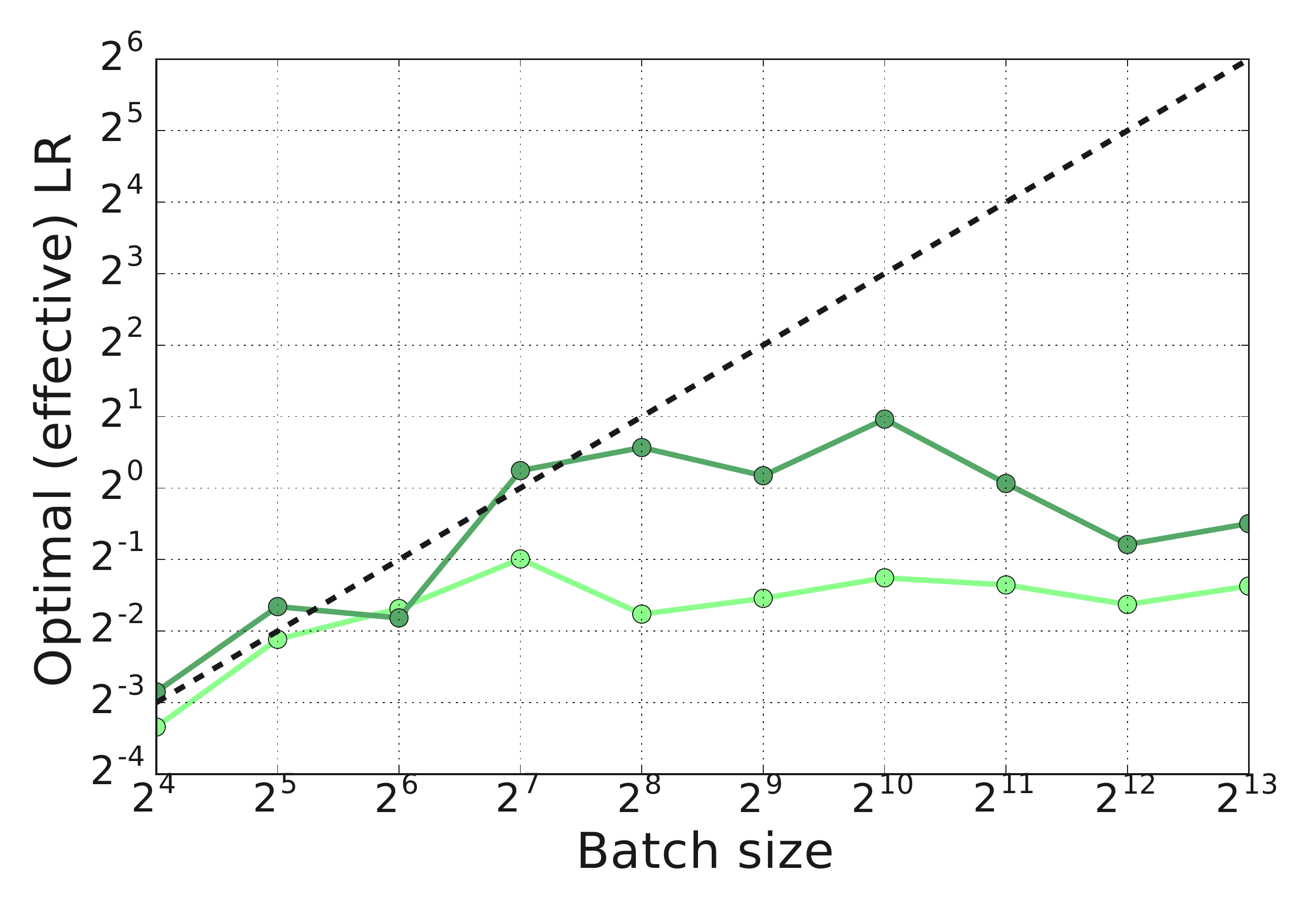}
    \end{subfigure}
    \begin{subfigure}[t]{0.295\textwidth}
        \centering
        \includegraphics[height=1.1in]{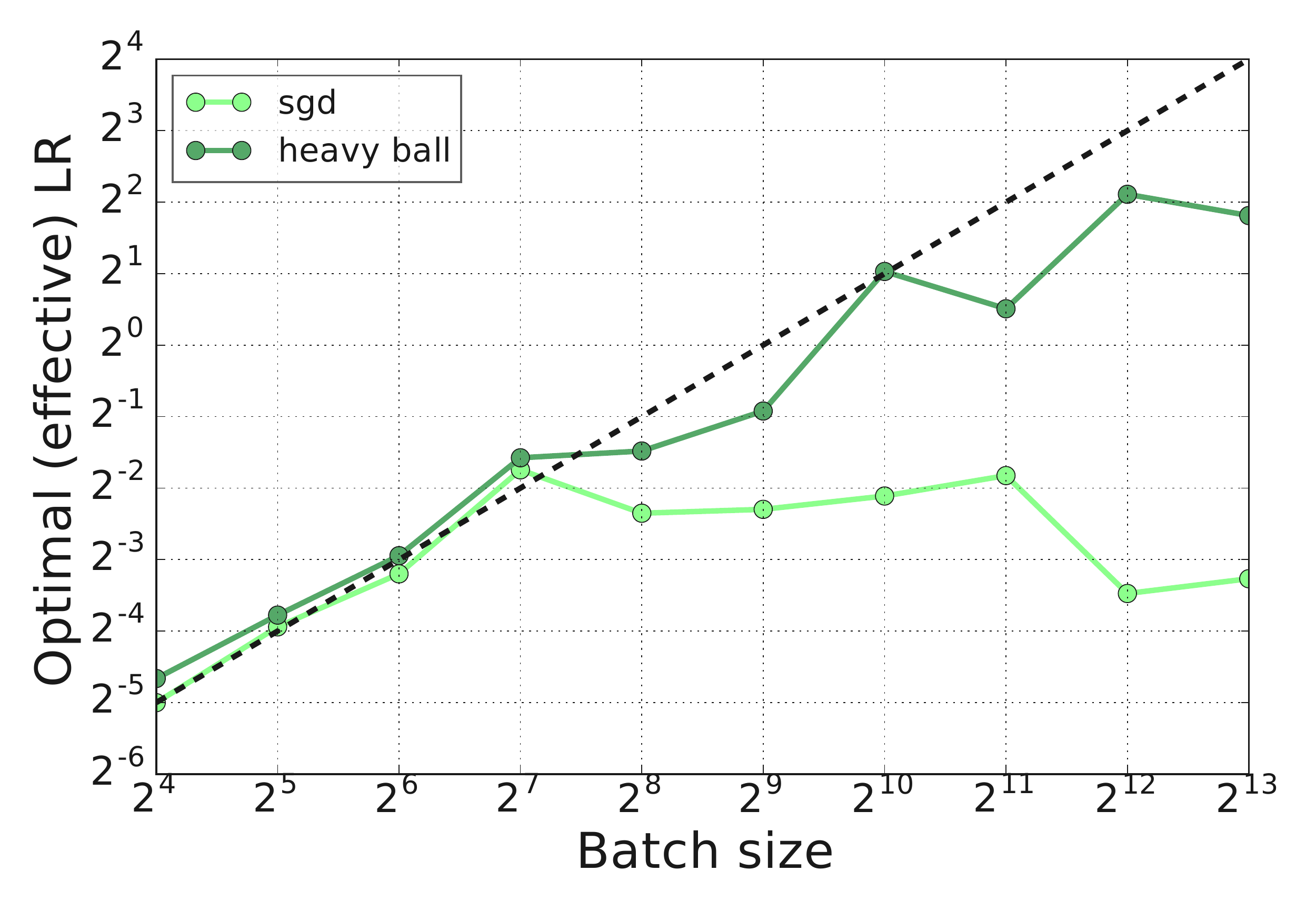}
    \end{subfigure}
    \vspace{-0.7cm}
    \caption{Optimal learning rates for plain SGD and momentum SGD. \textbf{Left}: Simple CNN on MNIST; \textbf{Right}: ResNet8 on CIFAR10}
    \label{fig:opt-lr}
\end{wrapfigure}
The NQM predicts that the optimal constant learning rate for plain SGD (or effective learning rate for momentum SGD) scales linearly with batch size initially, and then levels off after a certain batch size. Figure~\ref{fig:opt-lr} shows the empirical optimal (effective) learning rate as a function of batch size for simple CNN on MNIST and ResNet8 on CIFAR10. 
For small batch sizes, the optimal learning rate of plain SGD appears to match the optimal effective learning rate of momentum SGD. However, after a certain batch size, the optimal learning rate for plain SGD saturates while the optimal effective learning rate of momentum SGD keeps increasing. Interestingly, plain SGD and momentum SGD appear to deviate at the same batch size in the optimal effective learning rate and steps to target plots (Figures~\ref{fig:large-scale} and~\ref{fig:opt-lr}).

\subsection{Steps to Target on the Training Set}
\begin{wrapfigure}[10]{R}{0.6\textwidth}
	\vspace{-0.4cm}
    \centering
    \begin{subfigure}[t]{0.295\textwidth}
        \centering
        \includegraphics[height=1.1in]{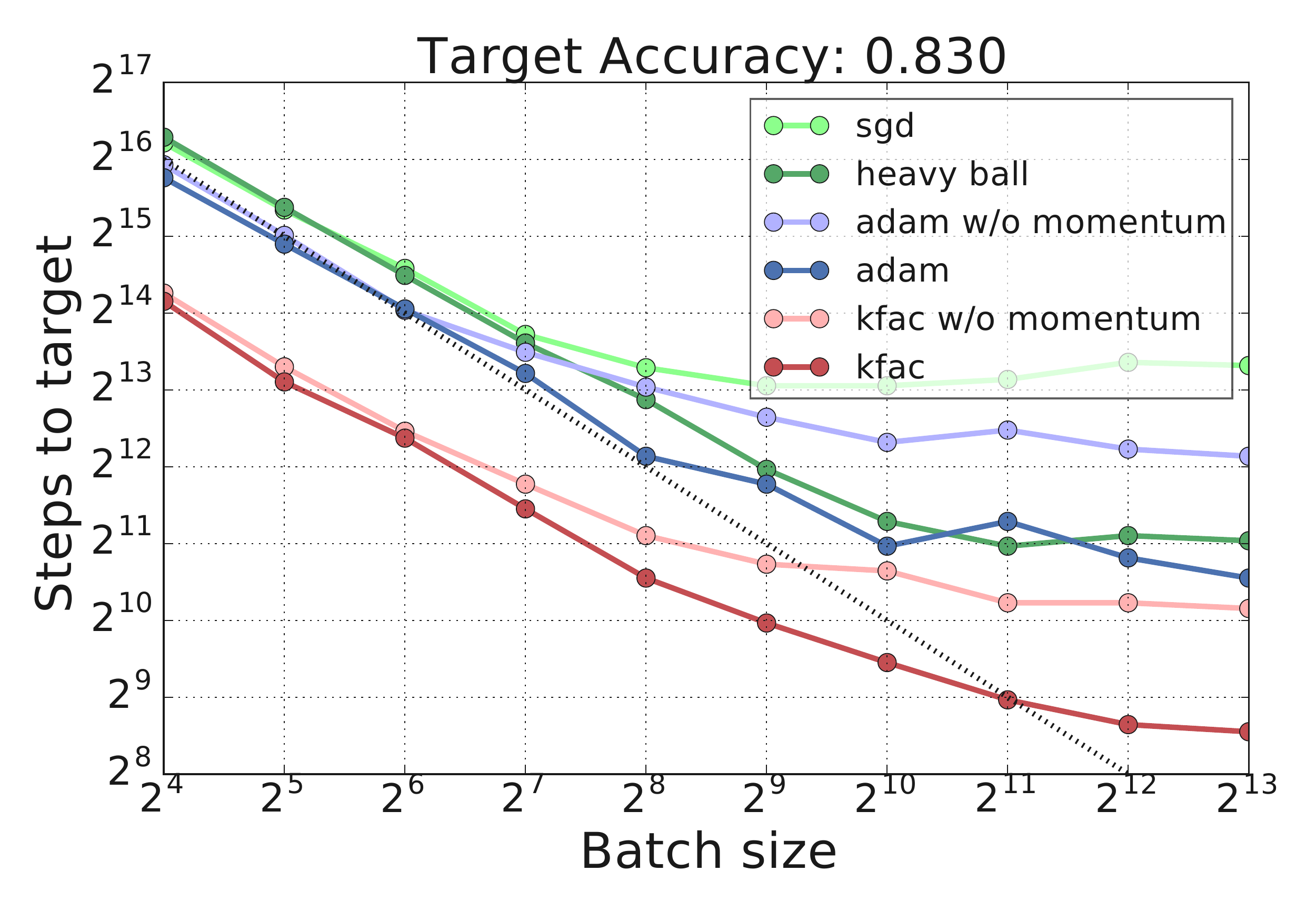}
    \end{subfigure}
    \begin{subfigure}[t]{0.295\textwidth}
        \centering
        \includegraphics[height=1.1in]{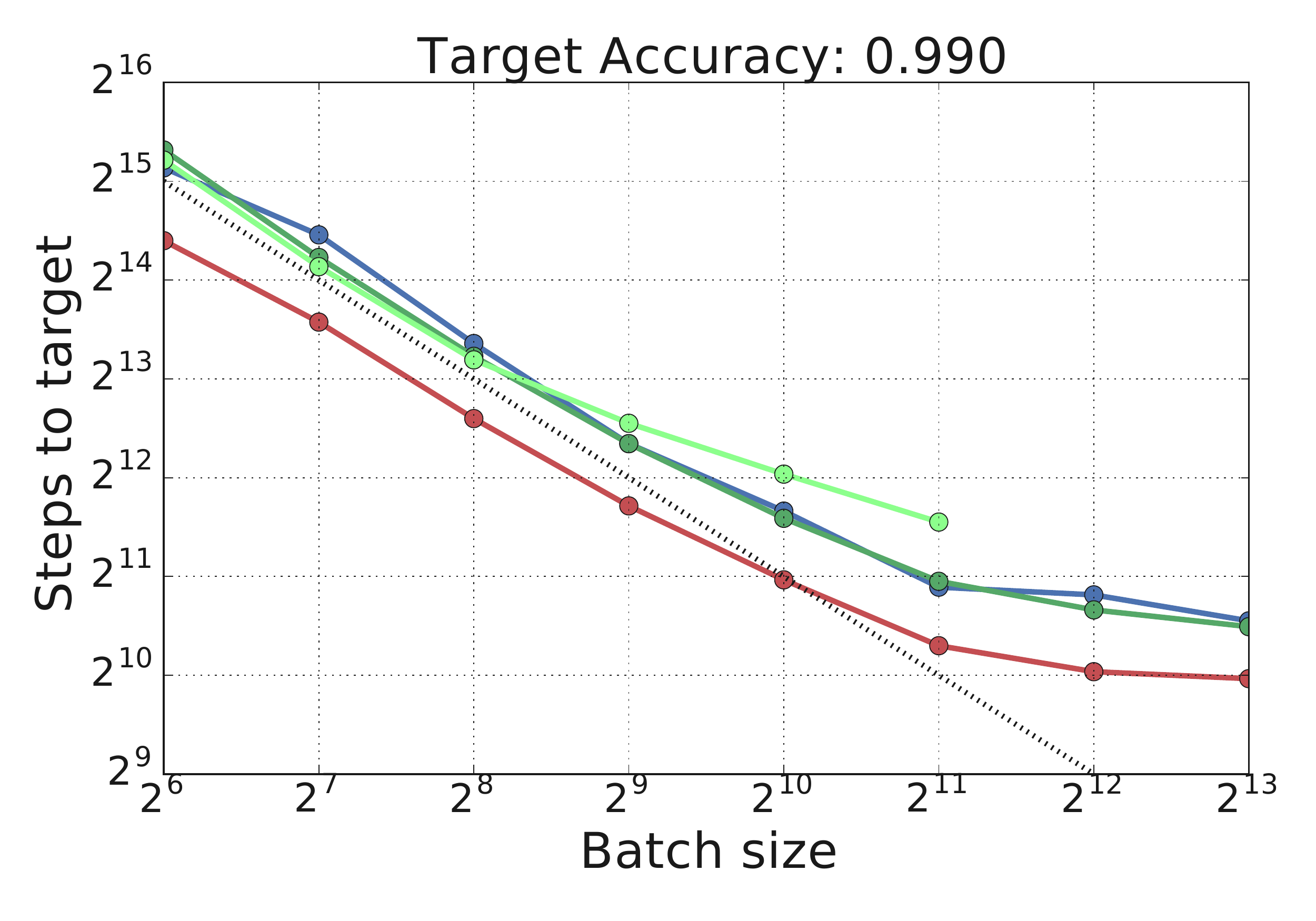}
    \end{subfigure}
    \vspace{-0.6cm}
    \caption{Steps to training accuracy versus batch size on CIFAR10. \textbf{Left:} ResNet8; \textbf{Right}: ResNet32.}
    \label{fig:training-target}
    \vspace{-0.4cm}
\end{wrapfigure}
Figure~\ref{fig:training-target} shows the empirical relationship between batch size and steps to target, measured on the training set, for ResNet8 and ResNet32 on CIFAR10. For ResNet8, the curves are almost identical to those using validation accuracy (Figure~\ref{subfig:resnet8-cifar}), but for ResNet32, the gaps between different optimizers become much smaller than in Figure~\ref{subfig:resnet32-cifar} and the effects of momentum and preconditioning appear to become less significant. Nevertheless, the qualitative differences between optimizers are consistent with the validation set measurements.

\section{Conclusion}

In this work, we analyzed the interactions between the batch size and the optimization algorithm from two perspectives: experiments with real neural networks, and a noisy quadratic model with parameters chosen based on empirical observations about neural networks. Despite its simplicity, the noisy quadratic model agrees remarkably well with a variety of neural network training phenomena, including learning rate scaling, critical batch sizes, and the effects of momentum, preconditioning and averaging. 
More importantly, the noisy quadratic model allows us to run experiments in seconds, while it can take weeks, or even months, to conduct careful large-scale experiments with real neural networks.
Therefore, the noisy quadratic model is a convenient and powerful way to quickly formulate testable predictions about neural network optimization.

\bibliography{neurips_2019.bib}
\bibliographystyle{plainnat}

\appendix
\newpage

\section{Kronecker-factored Approximate Curvature (K-FAC)}\label{app:kfac}
Kronecker-factored approximate curvature (K-FAC)~\citep{martens2015optimizing} uses a Kronecker-factored approximation to the curvature matrix to perform efficient approximate natural gradient updates. Considering the $l$-th layer in a neural network whose input activations are $\ba \in \mathbb{R}^{n}$, weight matrix $\weight \in \mathbb{R}^{n \times m}$, and outputs $\bs \in \mathbb{R}^{m}$, we have $\bs = \weight^\top \ba$. Therefore, the weight gradient is $\nabla_{\weight}\loss = \ba(\nabla_{\bs} \loss )^\top$. With this formula, K-FAC decouples this layer's Fisher matrix $\fisher$ using an independence assumption:
\begin{equation}
\begin{aligned}
	\fisher &= \expect[\newvec \{\nabla_{\weight} \loss\}\newvec\{\nabla_{\weight}\loss \}^\top] = \expect[\{\nabla_{\bs} \loss \}\{\nabla_{\bs} \loss\}^\top \kron \ba\ba^\top] \\
    &\approx \expect [\{\nabla_{\bs} \loss\}\{\nabla_{\bs} \loss \}^\top] \kron \expect[\ba\ba^\top] = \bS \kron \bA
\end{aligned}
\end{equation}
where $\bA = \expect [\ba\ba^\top]$ and 
$\bS = \expect [\{ \nabla_{\bs} \loss \} \{\nabla_{\bs} \loss \}^{\top}]$.
Decomposing $\fisher$ into $\bA$ and $\bS$ not only avoids the quadratic storage cost of the exact Fisher, but also enables tractable computation of the approximate natural gradient:
\begin{equation}\label{eq:kfac-inverse}
\begin{aligned}
	\fisher^{-1}\newvec\{\nabla_{\weight}\loss\} &= (\bS^{-1} \otimes \bA^{-1}) \, \newvec\{\nabla_{\weight}\loss\} \\ 
	&= \newvec[\bA^{-1} \nabla_{\weight}\loss \bS^{-1}]
\end{aligned}
\end{equation}
As shown by eqn.~\eqref{eq:kfac-inverse}, computing natural gradient using K-FAC only consists of matrix transformations comparable to size of $\weight$, making it very efficient. 

Later, \citet{grosse2016kronecker} further extended K-FAC to convolutional layers under additional assumptions of spatial homogeneity~(\textbf{SH}) and spatially uncorrelated derivatives~(\textbf{SUD}). Suppose the input $\ba \in \real^{c_{\mathrm{in}} \times h \times w}$ and the output $\bs \in \real^{c_{\mathrm{out}} \times h \times w}$, then the gradient of the reshaped weight $\weight \in \real^{c_{\mathrm{out}}\times c_{\mathrm{in}}k^2}$ is $\nabla_{\weight}\loss = \sum\ba_i\nabla_{\bs_{i}}\loss  ^\top$, and the corresponding Fisher matrix is:
\begin{equation}
\begin{aligned}
	\fisher &\approx \sum\expect\left[ \{\nabla_{\bs_{i}}\loss \}\{\nabla_{\bs_{i'}}\loss \}^\top\right] \kron \expect\left[\ba_{i}\ba_{i'}^\top\right] \\
	&\approx \underbrace{\left(\frac{1}{|\mathcal{I}|}\sum\expect\left[ \{\nabla_{\bs_{i}}\loss \}\{\nabla_{\bs_{i}}\loss \}^\top\right]\right)}_{\bS, \mathrm{size}=(c_{\mathrm{out}})^2} \kron \underbrace{\left(\sum\expect\left[\ba_{i}\ba_{i}^\top\right]\right)}_{\bA, \mathrm{size}=(c_{\mathrm{in}} \times k^2)^2}
\end{aligned}
\end{equation}
where $\mathcal{I} = [h]\times [w]$ is the set of spatial locations, $\ba_i \in \mathbb{R}^{c_{\mathrm{in}}k^2}$ is the patch extracted from $\ba$, $\nabla_{\bs_{i}}\loss \in \mathbb{R}^{c_{\mathrm{out}}}$ is the gradient to each spatial location in $\bs$ and $i, i' \in \mathcal{I}$. 

\subsection{K-FAC for Transformer}

K-FAC has been implemented on the autoencoder~\citep{martens2015optimizing} and various convolutional networks~\citep{grosse2016kronecker, ba2017distributed} before. To our knowledge, this is the first time K-FAC is implemented on the Transformer model. What is different from the previous models is the shared weight matrix between the embedding layer and the pre-softmax linear transformation~\citep{vaswani2017attention}. In particular, the weight matrix is transposed at the pre-softmax layer: $\bs = \weight \ba$ and $\nabla_{\weight}\loss = (\nabla_{\bs} \loss ) \ba^\top$. With the same assumptions as the non-transposed case, we get
\begin{equation}
    \fisher \approx \expect [\ba\ba^\top \kron \{\nabla_{\bs} \loss\}\{\nabla_{\bs} \loss \}^\top] = \bA \kron \bS
\end{equation}
i.e. the positions of the two Kronecker factors are swapped. If we name the two Kronecker factors "input factor" and "output factor" respectively, i.e. $\fisher \approx \textit{input\_factor} \kron \textit{output\_factor}$, then for the weight matrix that is shared between the embedding layer and the pre-softmax layer, the \textit{input\_factor} has contributions from both the embedding inputs and the gradients of pre-softmax layer outputs; and the \textit{output\_factor} has contributions from both the pre-softmax layer inputs and the gradients of the embedding outputs. In practice, when computing a Kronecker factor, we treat contribution from multiple sources as an equivalent situation as contribution from multiple training examples from a mini-batch. Also note that because of the high dimensionality of the embedding weight matrix (with a vocabulary size of 32,768), the dense input factor would have size $[32768, 32768]$. In order to save memory, we use a diagonal matrix to estimate the \textit{input\_factor}. The \textit{output\_factor} is still estimated with a dense matrix.

\section{Dynamics of momentum SGD on noisy quadratic model}
Similar to plain SGD, by treating $\theta_i$ as a random variable, we can explicitly write down the dynamics of its expectation and variance. But due to the use of momentum, we need to take into account $m_i$ and its correlation with $\theta_i$. Because each dimension evolves independently, we drop the the dimension subscripts. We first calculate the expectation of the parameter and velocity:
\begin{equation}
\begin{aligned}
    \expect \left[\theta(t+1) \right] &= (1 - \lr h) \expect\left[\theta(t) \right] - \lr \beta \expect\left[m(t) \right] \\
    \expect \left[m(t+1) \right] &= \beta \expect\left[m(t) \right] + h \expect\left[\theta(t) \right]
\end{aligned}
\end{equation}
We then calculate the variance:
\begin{equation}
\begin{aligned}
    \vars \left[\theta(t+1) \right] &= (1 - \lr h)^2 \vars\left[\theta(t) \right] + (\lr \beta)^2 \expect\left[m(t) \right] - 2(1 - \lr h)\lr \beta \cov(t) + \frac{\lr^2 c}{B} \\
    \vars \left[m(t+1) \right] &= \beta^2 \vars\left[m(t) \right] + h^2 \vars\left[\theta(t) \right] + 2\beta h \cov(t) + \frac{c}{B}
\end{aligned}
\end{equation}
where $\cov(t) = \cov(\theta(t), m(t))$ evolves as
\begin{equation}
    \cov(t+1) = (1 - \lr h) h \vars\left[ \theta(t) \right] - \lr \beta^2 \vars \left[m(t) \right] + (1 - 2\lr h) \beta \cov(t) - \frac{\lr c}{B}
\end{equation}
Because the expected risk is totally decided by $\expect\left[\theta \right]^2 + \vars \left [\theta \right]$, we define $A(\cdot) = \expect\left[\cdot \right]^2 + \vars \left[\cdot \right]$ and $C(t) = \expect[\theta(t)]\expect[m(t)] + \cov(\theta(t), m(t))$. We can then simplify the dynamics as follows
\begin{equation}\label{eq:transition}
\begin{aligned}
    & A(\theta(t+1)) = (1 - \lr h)^2 A(\theta(t)) + (\lr \beta)^2 A(m(t)) - 2(1 - \lr h) \lr \beta C(t) + \frac{\lr^2 c}{B} \\
    & A(m(t+1)) = \beta^2 A(m(t)) + h^2 A(\theta(t)) + 2\beta h C(t) + \frac{c}{B} \\
    & C(t+1) = (1 - \lr h) h A(\theta(t)) - \lr \beta^2 A(m(t)) + (1 - 2\lr h)\beta C(t) - \frac{\lr c}{B}
\end{aligned}
\end{equation}
or equivalently
\begin{equation}\label{eq:transition-2}
\underbrace{\begin{bmatrix}
A(\theta(t+1)) \\
\lr^2 A(m(t+1)) \\
-\lr C(t+1) \\
\end{bmatrix}}_{{\mathbf{v}(t+1)}}
=
\underbrace{
\begin{bmatrix}
(1 - \lr h)^2 & \beta^2 & 2(1 - \lr h)\beta \\
(\lr h)^2 & \beta^2 & -2 \beta \lr h \\
-(1 - \lr h)\lr h & \beta^2 & (1 - 2\lr h)\beta
\end{bmatrix}}_{\text{transition matrix} \; \mathbf{T}}
\underbrace{\begin{bmatrix}
A(\theta(t)) \\
\lr^2 A(m(t)) \\
-\lr C(t) \\
\end{bmatrix}}_{{\mathbf{v}(t)}}
+
\underbrace{\begin{bmatrix}
\frac{\lr^2 c}{B} \\
\frac{\lr^2 c}{B} \\
\frac{\lr^2 c}{B} \\
\end{bmatrix}}_{\mathbf{n}}
\end{equation}
The convergence rate is determined by the transition matrix $\mathbf{T}$ which has the characteristic polynomial
\begin{equation}
    \left|\mathbf{T} - \lambda \iden \right| = - (\lambda - \beta) (\lambda^2 - (\beta^2 - 2\lr h \beta + (1 - \lr h)^2) \lambda + \beta^2)
\end{equation}
With the momentum value $\beta = (1 - \sqrt{\lr h})^2$, all eigenvalues of the transition matrix are equal to each other with the value $\beta$, giving the fastest convergence.

\section{Proof of Theorem~\ref{thm:dynamics}}
\label{app:momentum_proof}
For a linear dynamical system like eqn.~\eqref{eq:transition-2}, we can get $\mathbf{v}(t)$ in the following form:
\begin{equation}
\mathbf{v}(t) = \mathbf{T}^t \mathbf{v}(0) + \sum_{p=1}^{t+1} \mathbf{T}^{p-1} \mathbf{n} \leq \mathbf{T}^t \mathbf{v}(0) + \sum_{p=1}^{\infty} \mathbf{T}^{p-1} \mathbf{n}
\end{equation}
We first analyze the stochastic term $\sum_{p=1}^{\infty} \mathbf{T}^{p-1} \mathbf{n}$. For notational convenience, we define
\begin{equation}\label{eq:simplified-dyn}
    \sum_{p=1}^{\infty} \mathbf{T}^{p-1} \mathbf{n} \triangleq \sum_{p=0}^\infty \left[x_p, y_p, z_p \right]^\top
\end{equation}
In eqn.~\eqref{eq:simplified-dyn}, we append zero vector $[x_0, y_0, z_0]^\top$ for convenience. To compute the infinite sum, we first focus on a single term. We have the following update:
\begin{equation}
\begin{aligned}
    \sqrt{x_{p+1}} &= (1 - \lr h) \sqrt{x_{p}} + \beta \sqrt{y_p} \\
    \sqrt{y_{p+1}} &= -\lr h \sqrt{x_{p}} + \beta \sqrt{y_p} \\
\end{aligned}
\end{equation}
Since we only care $x_p$ which totally decide the loss, so we get rid of $y_p$ by merging two updates, which yields a second-order difference equation:
\begin{equation}
    \sqrt{x_{p+1}} = (1 - \lr h + \beta) \sqrt{x_{p}} - \beta \sqrt{x_{p-1}}
\end{equation}
with initial conditions $\sqrt{x_0} = 0$ and $\sqrt{x_1} = \sqrt{\frac{\lr^2 c}{B}}$. To solve the second-order difference equation, we leverage the Z-transform to get the analytical form. Based on basic manipulation of the Z-transform, we have the Z-domain function
\begin{equation}
    X(Z) = \frac{\sqrt{\frac{\lr^2 c}{B}} Z}{Z^2 - (1 - \lr h + \beta) Z + \beta} = \frac{\sqrt{\frac{\lr^2 c}{B}}}{r_1 - r_2}\left(\frac{1}{1 - Z^{-1} r_1} - \frac{1}{1 - Z^{-1} r_2} \right)
\end{equation}
where $r_1$ and $r_2$ are two roots of equation $z^2 - (1 - \lr h + \beta)z + \beta$. Then, we use the inverse Z-transform to get $\sqrt{x_p}$:
\begin{equation}
    \sqrt{x_p} = \sqrt{\frac{\lr^2 c}{B}}\frac{r_1^p - r_2^p}{r_1 - r_2}
\end{equation}
and therefore
\begin{equation}
    x_p = \frac{\lr^2 c}{B}\frac{r_1^{2p} + r_2^{2p} - 2(r_1 r_2)^p}{(r_1 - r_2)^2}
\end{equation}
Now, we are ready to compute the infinite sum $\sum_{p=0}^\infty x_p$:
\begin{equation}
\begin{aligned}
    \sum_{p=0}^\infty x_p &= \frac{\frac{\lr^2 c}{B}}{(r_1 - r_2)^2} \left(\frac{1}{1 - r_1^{2}} + \frac{1}{1 - r_2^2} - \frac{2}{1 - r_1 r_2} \right) \\
    &= \frac{\lr^2 c}{B}\frac{1 + r_1 r_2}{(1 - r_1^2)(1 - r_2^2)(1 - r_1 r_2)}
\end{aligned}
\end{equation}
Because $r_1$ and $r_2$ are two roots with $r_1 r_2 = \beta$, $r_1 + r_2 = 1 - \lr h + \beta$, we have
\begin{equation}\label{eq:ssk-momentum}
    \sum_{p=0}^\infty x_p = \frac{\lr c (1 + \beta)}{B h (2\beta + 2 - \lr h) (1 - \beta)}
\end{equation}
Now, we analyze the deterministic term. 
Similar to the analysis of stochastic term, we have the same second-order difference equation
\begin{equation}
    \sqrt{x^\prime_{p+1}} = (1 - \lr h + \beta) \sqrt{x^\prime_{p}} - \beta \sqrt{x^\prime_{p-1}}
\end{equation}
except the initial conditions become $\sqrt{x^\prime_0} = \sqrt{x^\prime_1} = \sqrt{A(\theta(0))}$. According to Z-transform, we have
\begin{equation}
    x^\prime_t= \left(\frac{r_1^{t+1} - r_2^{t+1} - \beta (r_1^{t} - r_2^{t})}{r_1 - r_2} \right)^2A(\theta(0)) 
\end{equation}
Along with eqn.~\eqref{eq:ssk-momentum}, we have
\begin{equation}
    A(\theta(t)) \leq \left(\frac{r_1^{t+1} - r_2^{t+1} - \beta (r_1^{t} - r_2^{t})}{r_1 - r_2} \right)^2A(\theta(0)) + \frac{\lr c (1 + \beta)}{B h (2\beta + 2 - \lr h) (1 - \beta)}
\end{equation}

\section{Proof of Theorem~\ref{thm:ema-dynamics}}
Similar to plain SGD, by treating $\theta_i$ as a random variable, we can explicitly write down the dynamics of its expectation and variance. But due to the use of moving averaging, we need to take into account $\tilde{\theta}_i$ and its correlation with $\theta_i$. Because each dimension evolves independently, we drop the the dimension subscripts. We first calculate the expectation of the parameter and the average:
\begin{equation}
\begin{aligned}
    \expect \left[\theta(t+1) \right] &= (1 - \lr h) \expect\left[\theta(t) \right] \\
    \expect [\tilde{\theta}(t+1) ] &= \gamma \expect[\tilde{\theta}(t) ] + (1-\gamma)(1-\lr h)\expect \left[\theta(t) \right]
\end{aligned}
\end{equation}
We then calculate the variance:
\begin{equation}
\begin{aligned}
    \vars \left[\theta(t+1) \right] &= (1 - \lr h)^2 \vars\left[\theta(t) \right] + \frac{\lr^2 c}{B} \\
    \vars [\tilde{\theta}(t+1) ] &= \gamma^2 \vars[\tilde{\theta}(t) ] + (1 - \gamma)^2 (1 - \lr h)^2 \vars\left[\theta(t) \right] \\ 
    &+ 2\gamma (1 - \gamma)(1 - \lr h) \cov(t) + \frac{(1 - \gamma)^2\lr^2 c}{B}
\end{aligned}
\end{equation}
where $\cov(t) = \cov(\theta(t), \tilde{\theta}(t))$ evolves as
\begin{equation}
    \cov(t+1) = (1 - \gamma)(1 - \lr h)^2 \vars\left[\theta(t) \right] + (1 - \lr \gamma) \cov(t) + \frac{(1 - \gamma)\lr^2 c}{B}
\end{equation}
Because the expected risk is totally decided by $\expect[\tilde{\theta} ]^2 + \vars  [\tilde{\theta} ]$, we define $A(\cdot) = \expect\left[\cdot \right]^2 + \vars \left[\cdot \right]$ and $C(t) = \expect[\theta(t)]\expect[\tilde{\theta}(t)] + \cov(\theta(t), \tilde{\theta}(t))$. We can then simplify the dynamics as follows
\begin{equation}
\underbrace{
\begin{bmatrix}
A(\theta(t+1)) \\
\frac{A(\tilde{\theta}(t+1))}{(1-\gamma)^2} \\
\frac{C(t+1)}{(1-\gamma)} \\
\end{bmatrix}}_{\mathbf{v}(t+1)}
=
\underbrace{
\begin{bmatrix}
(1 - \lr h)^2 & 0 & 0 \\
(1 - \lr h)^2 & \gamma^2 & 2\gamma (1 - \gamma)(1 - \lr h) \\
(1 - \lr h)^2 & 0 & \gamma (1 - \lr h)
\end{bmatrix}}_{\text{transition matrix} \; \mathbf{T}}
\underbrace{
\begin{bmatrix}
A(\theta(t)) \\
\frac{A(\tilde{\theta}(t))}{(1-\gamma)^2} \\
\frac{C(t)}{(1-\gamma)} \\
\end{bmatrix}}_{\mathbf{v}(t)}
+
\underbrace{
\begin{bmatrix}
\frac{\lr^2 c}{B} \\
\frac{\lr^2 c}{B} \\
\frac{\lr^2 c}{B} \\
\end{bmatrix}}_{\mathbf{n}}
\end{equation}
For such a linear dynamical system, we can easily get the $\mathbf{v}(t)$ in the following form:
\begin{equation}
    \mathbf{v}(t) = \mathbf{T}^t \mathbf{v}(0) + \sum_{p=1}^{t+1} \mathbf{T}^{p-1} \mathbf{n} \leq \mathbf{T}^t \mathbf{v}(0) + \sum_{p=1}^{\infty} \mathbf{T}^{p-1} \mathbf{n}
\end{equation}
Now, to get the closed-form of $\mathbf{v}(t)$, we first analyze the second term which involves the infinite sum. For notational convenience, we introduce the following notations:
\begin{equation}\label{eq:sto-term}
    \sum_{p=1}^{\infty} \mathbf{T}^{p-1} \mathbf{n} \triangleq \sum_{p=0}^\infty \left[x_p, y_p, z_p \right]^\top
\end{equation}
In eqn.~\eqref{eq:sto-term}, we append zero vector $[x_0, y_0, z_0]^\top$ for convenience. To compute the infinite sum, we first focus on a single term. We have the following update:
\begin{equation}
\begin{aligned}
    \sqrt{x_{p+1}} &= (1 - \lr h) \sqrt{x_{p}}\\
    \sqrt{y_{p+1}} &= (1 - \lr h) \sqrt{x_{p}} + \gamma \sqrt{y_{p}} \\
\end{aligned}
\end{equation}
Since we only care $y_p$ which totally decide the loss, so we get rid of $x_p$ by merging two updates, which yields a second-order difference equation:
\begin{equation}\label{eq:ema-second}
    \sqrt{y_{p+1}} = (1 - \lr h + \gamma) \sqrt{y_{p}} - (1 - \lr h)\gamma \sqrt{y_{p-1}}
\end{equation}
with initial conditions $\sqrt{y_0} = 0$ and $\sqrt{y_1} = \sqrt{\frac{\lr^2 c}{B}}$. To solve the second-order difference equation, we leverage the Z-transform to get the analytical form. Based on basic manipulation of the Z-transform, we have the Z-domain function
\begin{equation}
    Y(Z) = \frac{\sqrt{\frac{\lr^2 c}{B}} Z}{Z^2 - (1 - \lr h + \gamma) Z + \gamma} = \frac{\sqrt{\frac{\lr^2 c}{B}}}{r_1 - r_2}\left(\frac{1}{1 - Z^{-1} r_1} - \frac{1}{1 - Z^{-1} r_2} \right)
\end{equation}
where $r_1$ and $r_2$ are two roots of equation $z^2 - (1 - \lr h + \gamma)z + (1-\lr h)\gamma$. Then, we use the inverse Z-transform to get $\sqrt{y_p}$:
\begin{equation}
    \sqrt{y_p} = \sqrt{\frac{\lr^2 c}{B}}\frac{r_1^p - r_2^p}{r_1 - r_2}
\end{equation}
and therefore
\begin{equation}
    y_p = \frac{\lr^2 c}{B}\frac{r_1^{2p} + r_2^{2p} - 2(r_1 r_2)^p}{(r_1 - r_2)^2}
\end{equation}
Now, we are ready to compute the infinite sum $\sum_{p=0}^\infty y_p$:
\begin{equation}\label{eq:random-eq}
\begin{aligned}
    \sum_{p=0}^\infty y_p &= \frac{\frac{\lr^2 c}{B}}{(r_1 - r_2)^2} \left(\frac{1}{1 - r_1^{2}} + \frac{1}{1 - r_2^2} - \frac{2}{1 - r_1 r_2} \right) \\
    &= \frac{\lr^2 c}{B}\frac{1 + r_1 r_2}{(1 - r_1^2)(1 - r_2^2)(1 - r_1 r_2)}
\end{aligned}
\end{equation}
It is easy to see that $r_1 = 1 - \lr h$ and $r_2 = \gamma$, we then plug them back into eqn.~\eqref{eq:random-eq} and get
\begin{equation}
    \sum_{p=0}^\infty y_p = \frac{\lr c (1 + (1 - \lr h)\gamma)}{B h (2 - \lr h) (1 - \gamma^2) (1 - (1 - \lr h)\gamma)}
\end{equation}
For the other term $\mathbf{T}^t \mathbf{v}(0)$, we can reuse the same second-order difference equation~\eqref{eq:ema-second} except with initial conditions $\sqrt{y_0} = \sqrt{y_1} = \frac{1}{1 - \gamma}\sqrt{A(\theta(0))}$. According to Z-transform, we have
\begin{equation}
    y_t = \frac{1}{(1 - \gamma)^2}\left(\frac{(r_1^{t+1} - r_2^{t+1}) - \gamma(1 - \lr h)(r_1^t - r_2^t)}{r_1 - r_2} \right)^2 A(\theta(0))
\end{equation}
Therefore, we have the following upper bound:
\begin{equation}
    A(\tilde{\theta}(t)) \leq \left(\frac{r_1^{t+1} - r_2^{t+1} - \gamma(1 - \lr h)(r_1^t - r_2^t)}{r_1 - r_2} \right)^2 A(\theta(0)) + \frac{\lr c (1 - \gamma)(1 + (1 - \lr h)\gamma)}{B h (2 - \lr h) (1 + \gamma) (1 - (1 - \lr h)\gamma)}
\end{equation}

\section{More results on the NQM}
\subsection{Eigenspectra of Neural Networks}
\begin{wrapfigure}[14]{R}{0.4\textwidth}
    \centering
    \vspace{-0.4cm}
    \includegraphics[width=2.1in]{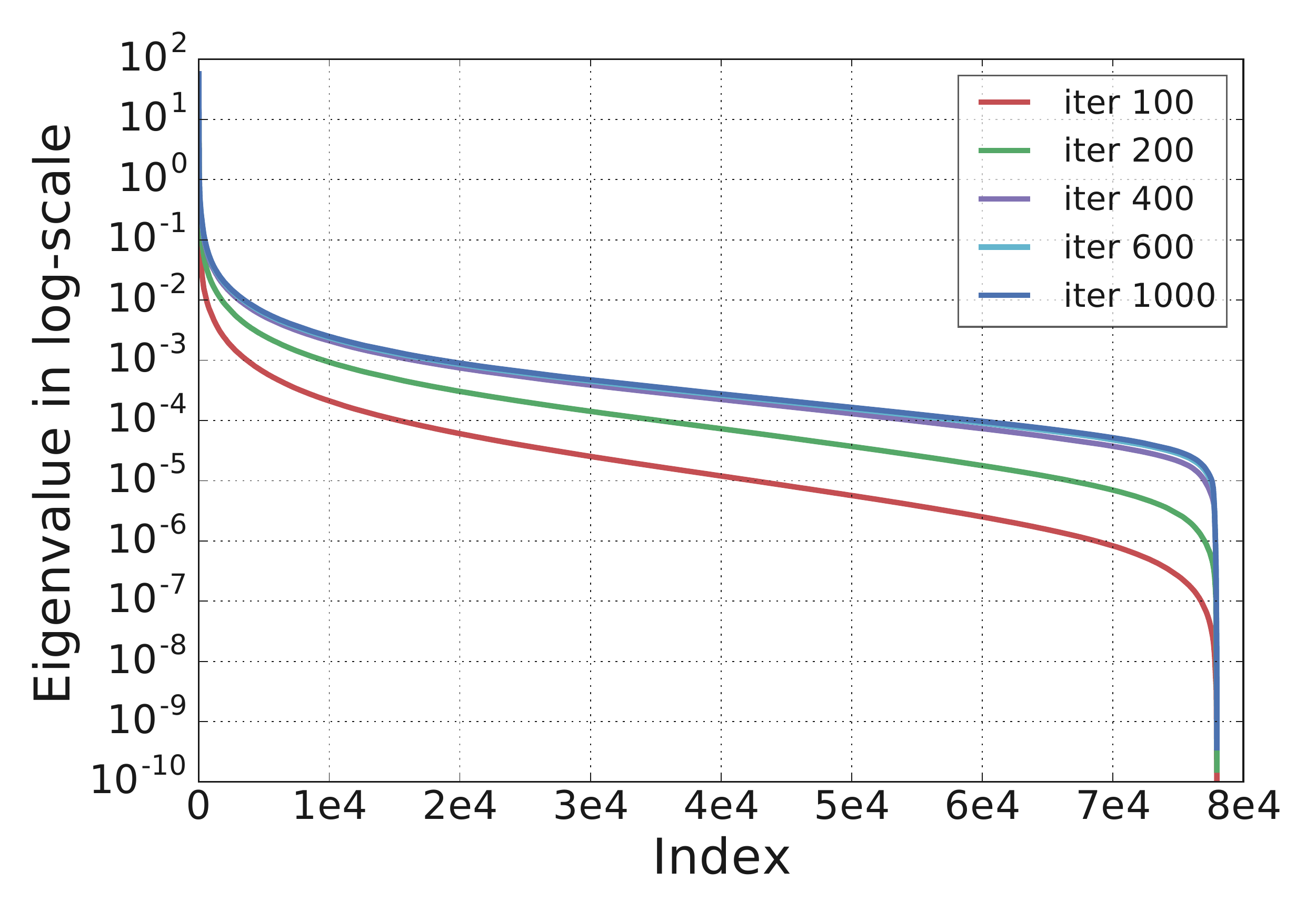}
    \vspace{-0.2cm}
    \caption{Eigenspectra of the K-FAC approximate Fisher matrix of ResNet8 at different training iterations. The model is trained on CIFAR-10 with batch size 3000.}
    \label{fig:eigenspectra}
    \vspace{-0.4cm}
\end{wrapfigure}
The main objective of this section is to examine the loss surface of modern neural networks in different stages of training in order to justify the assumptions made in NQM. Nevertheless, it is hard to visualize such a high dimensional space. Following recent work~\citep{sagun2016eigenvalues, ghorbani2019investigation}, we instead focus on analyzing the eigenspectrum of the Hessian/Fisher matrices. The Hessian/Fisher of the training loss (with respect to the parameters) is crucial in determining many behaviors of neural networks. The eigenvalues of the Hessian/Fisher characterize the local curvature of the loss surface which determines many training behaviors, including first-order methods optimization rates (at least for convex problems.)

It has been noted that the \emph{true} Fisher matrix is equivalent to the generalized Gauss-Newton Hessian matrix~\citep{martens2014new}, so we take it as a proxy of the Hessian.
To construct the eigenspectrum of the true Fisher matrix, we first leverage the Kronecker-factored approximation of the Fisher to get an estimation of the eigenspectrum, which may shed light upon the true eigenspectrum. Specifically, we train the network with K-FAC and then perform eigen-decomposition on saved Kronecker factors of the Fisher to calculate the eigenvalues.

The eigenspectra are plotted in Figure~\ref{fig:eigenspectra}. One interesting observation is that there are only a few large eigenvalues and a few small eigenvalues in the approximate Fisher matrices; the bulk of eigenvalues are in the middle of the spectrum. We also note that after 200 iterations of training the eigenspectrum remains mostly unchanged.

\begin{figure}[h]
\centering
\begin{subfigure}[t]{0.95\textwidth}
    \centering
    \includegraphics[height=1.25in]{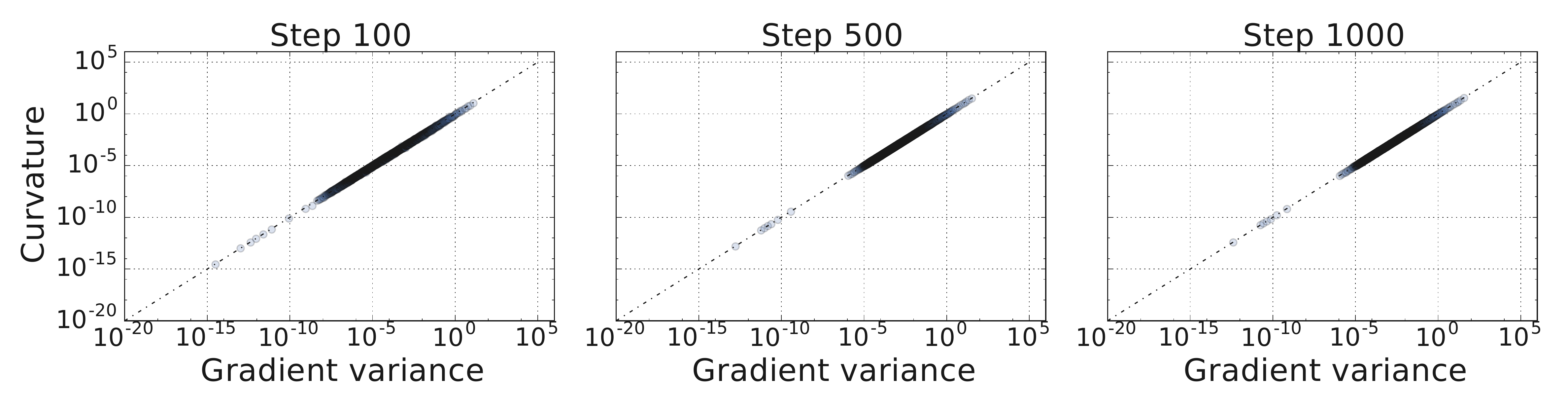}
    \vspace{-0.2cm}
    \caption{ResNet8}
    \label{subfig:plain-sgd-optlr}
\end{subfigure}
\begin{subfigure}[t]{0.95\textwidth}
    \centering
    \includegraphics[height=1.25in]{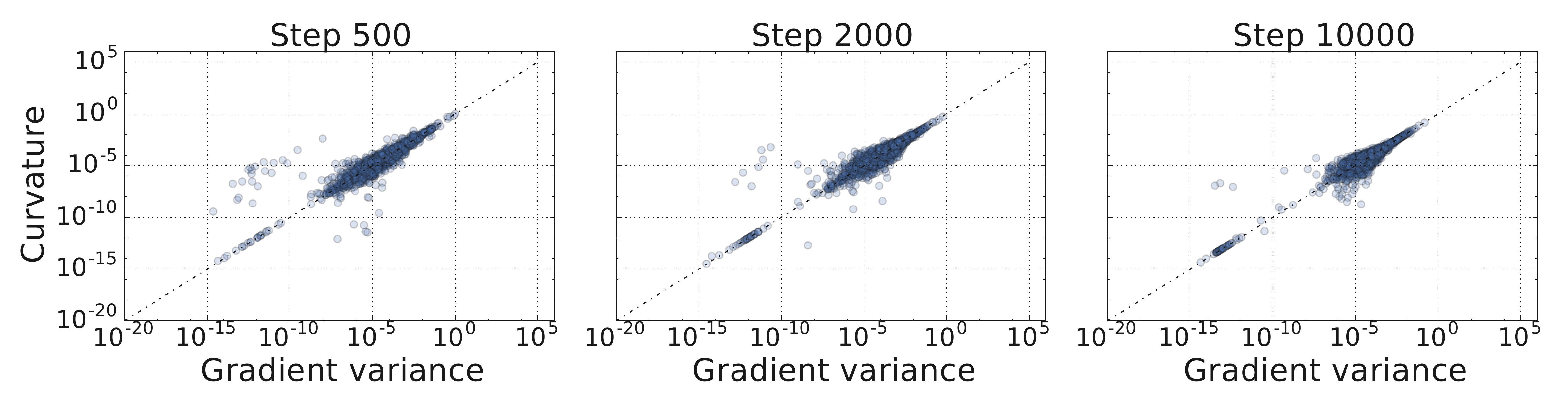}
    \vspace{-0.2cm}
    \caption{Transformer}
\end{subfigure}
\caption{\textbf{Scatter plots of second moment v.s. variance of gradients.} The gradients are projected onto the Kronecker-factored eigenbasis, which approximates the eigenbasis of the true Fisher. Each point compares the gradient variance and the second moment of the gradient in the direction of an eigenvector of the K-FAC approximated Fisher.}
\vspace{-0.3cm}
\label{fig:grad-var} 
\end{figure}
\subsection{Gradient Covariance in the Kronecker-Factored Eigenbasis}
\label{app:grad-var}
To verify the assumption in Section~\ref{sec:hessian-covariance} that $\mathbf{H}$ and $\mathbf{C}$ are codiagonalizable, we test it on practical neural networks by comparing the gradient variance to the curvature. This assumption is motivated by theoretical considerations that suggest $\mathbf{H} \approx \mathbf{C}$ for neural network training~\citep{martens2014new}. Ideally, we would like to compare the gradient variance and the curvature of the Fisher in the directions of the eigenvectors of the true Fisher. However, it is typically infeasible to get all these eigenvectors, especially for low curvature directions. To resolve this we instead use the Kronecker-factored eigenbasis~\citep{george2018fast, bae2018eigenvalue, wang2019eigendamage}, which is obtained from the K-FAC approximation. For this experiment, we are not relying on this basis being an accurate approximation to the eigendecomposition of the true Fisher; rather, we use the eigenbasis only as a way to obtain a diverse set of directions with both high and low curvature. For a given eigenvector $\mathbf{v}$, we project the gradients $\mathbf{g}$ of each training example onto $\mathbf{v}$ and compute the gradient variance $\cov(\mathbf{v}^\top \mathbf{g})$, as well as the curvature $\mathbf{v}^\top \fisher \mathbf{v}$. (The latter quantity can be obtained using matrix-vector products \citep{schraudolph2002fast}.)
As shown in Figure~\ref{fig:grad-var}, the gradient variances closely match the curvature (especially for the ResNet8 model on CIFAR10), validating our assumption that $\hessian \approx \covariance$.

\subsection{Plots for the Evolution of the First Term in Eqn.~\eqref{eq:dynamics}}\label{app:deterministc-term}
\begin{figure}[h]
	\vspace{-0.2cm}
    \centering
    \includegraphics[width=\textwidth]{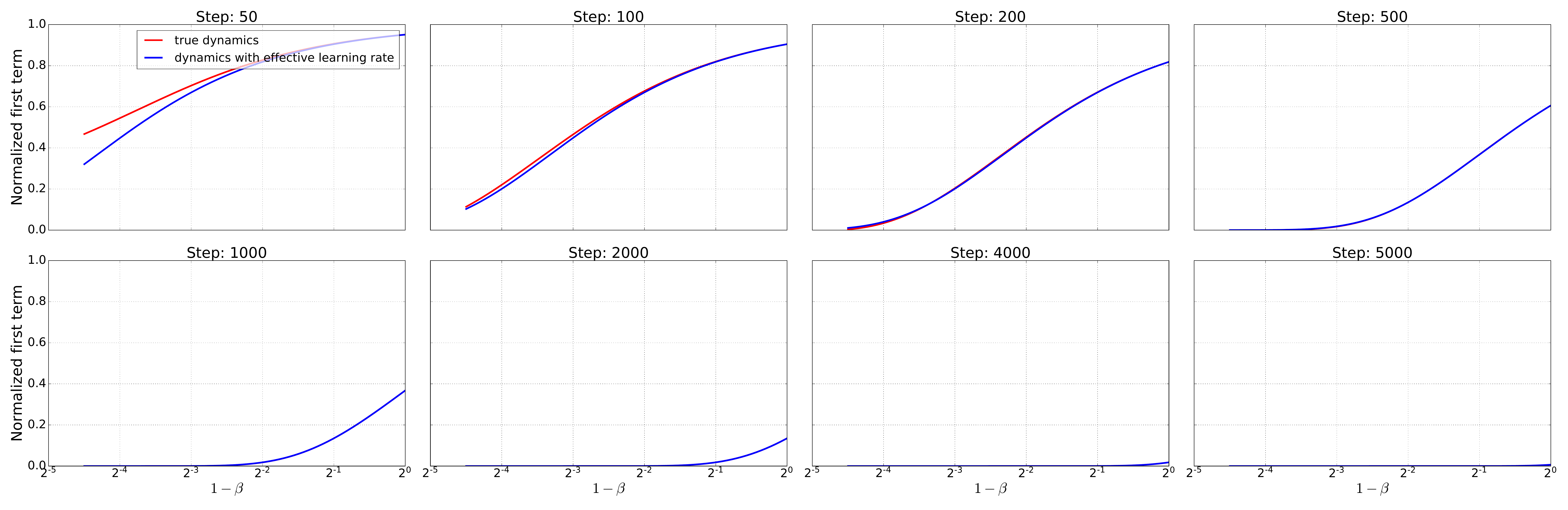}
    \vspace{-0.5cm}
    \caption{\textbf{Comparison in convergence between momentum SGD and SGD with adjusted learning rate.}  This plot shows values for the first term in eqn.~\eqref{eq:dynamics} as a function of $(1-\beta)$, which is the scaling between the ``effective learning rate'' and the true learning rate for momentum SGD. The red curves show the first term when using momentum, while the blue curves show the first term when using plain SGD with the learning rate set to the effective learning rate of momentum.}
    \label{fig:determinsitic term}
\end{figure}
In Section~\ref{sec:role-momentum}, we claim that the convergence of momentum SGD for a single dimension is very close to that of plain SGD with an adjusted learning rate (note that we already verified that the steady state risk of momentum SGD matches plain SGD using effective learning rate in Figure~\ref{fig:momentum}). Here we verify this argument by comparing them in the NQM. The total risk consists of two terms (eqn.~\eqref{eq:dynamics}): the first term determines convergence, while the second term (steady state risk) stays constant throughout training. Given that the second stays unchanged, we only plot the first term of eqn.~\eqref{eq:dynamics} in Figure~\ref{fig:determinsitic term}. Note that the values are normalized in the figures. We observe that the convergence dynamics of the two update rules closely match each other. For this experiment we set $\alpha h = 0.0005$, but the results are not sensitive to this value.

\subsection{Verification of Eigenspectrum}
\begin{figure}[h]
	\vspace{-0.2cm}
    \centering
    \includegraphics[width=\textwidth]{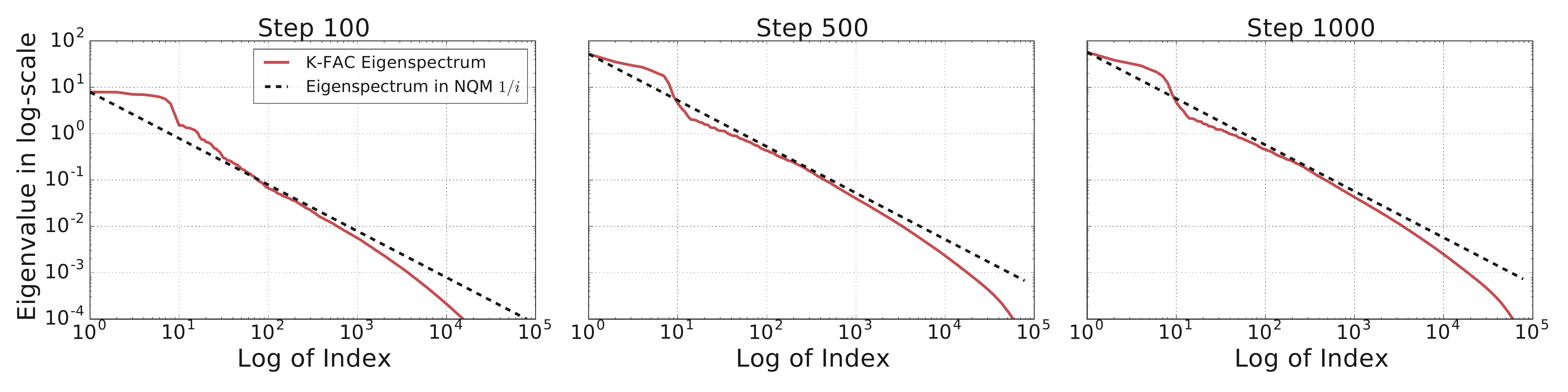}
    \vspace{-0.5cm}
    \caption{\textbf{Comparison between K-FAC Fisher eigenspectra and the $\frac{1}{i}$ distribution used in the NQM.}}
    \label{fig:eigenspectra-log}
\end{figure}

In Section~\ref{sec:nqm-exp}, we assume the diagonal entries of $\hessian$ are $\{\frac{1}{i}\}_{i=1}^d$. To justify this choice, we compare the K-FAC eigenspectra of ResNet8 to this distribution in Figure~\ref{fig:eigenspectra-log}. The distribution of eigenvalues we chose for $\hessian$ in the NQM very closely matches the eigenspectra of the real neural network, validating the assumption that the diagonal entries of $\hessian$ are $\{\frac{1}{i} \}_{i=1}^d$ in Section~\ref{sec:hessian-covariance}.
\label{app:verification-eigen}

\subsection{Effect of Loss Threshold}\label{app:loss-thres}
\begin{wrapfigure}[12]{R}{0.39\textwidth}
	\vspace{-0.4cm}
    \centering
    \includegraphics[width=1.9in]{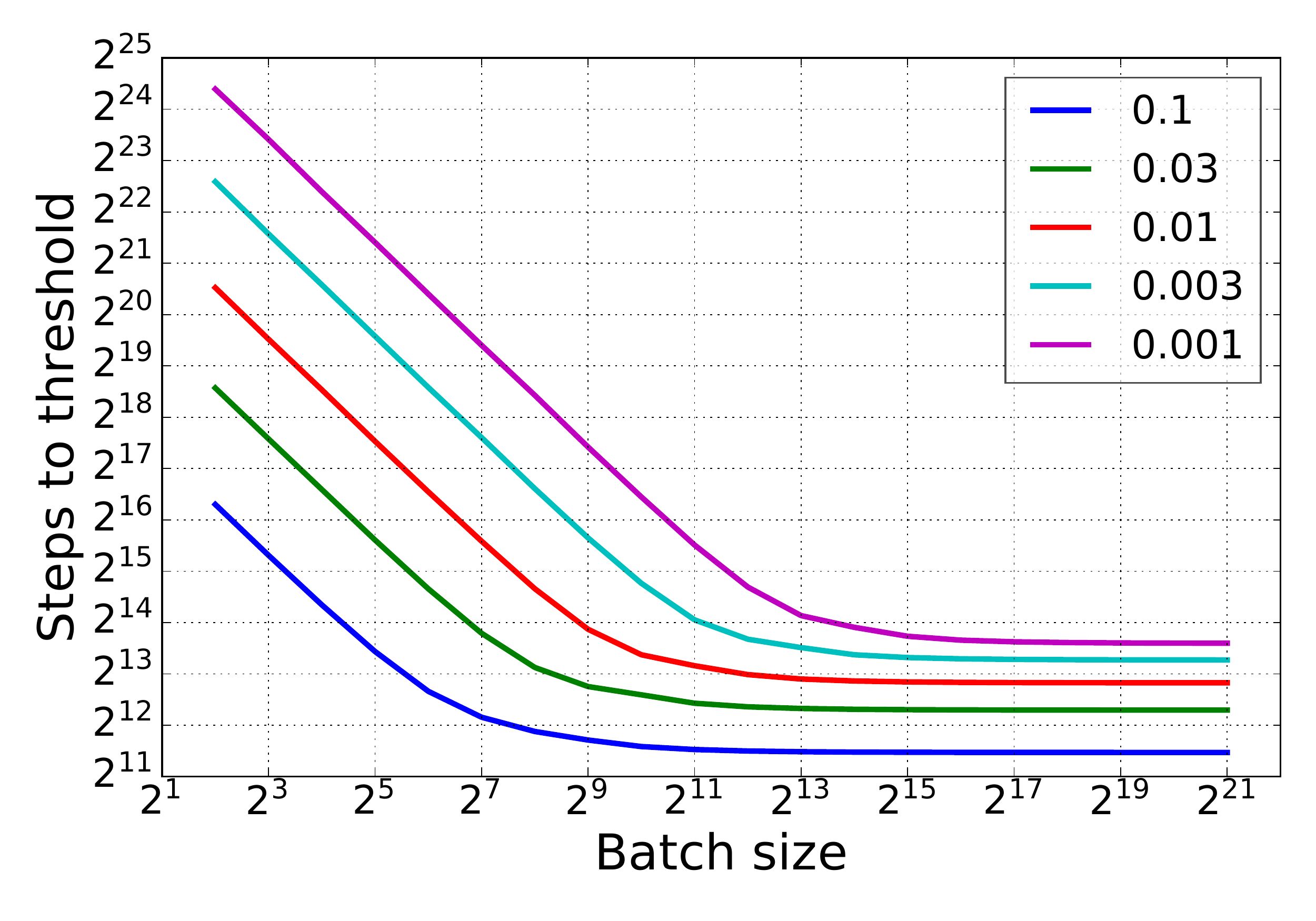}
    \vspace{-0.4cm}
    \caption{Number of training steps required to reach a target loss as a function of batch size for different loss threshold values.}
    \label{fig:loss_threshold}
\end{wrapfigure}
Recall that a main objective of this work is to characterize the effects of increasing the batch size on training time, as measured in the number of steps necessary to reach a goal target error/loss. Here we experiment with different loss thresholds to study the relationship between batch size and number of training steps. To obtain the minimal training steps for a given batch size, we do grid search over constant learning rates. Figure~\ref{fig:loss_threshold} shows that increasing the batch size initially decreases the required number of training steps proportionally, but eventually there are diminishing returns, which matches the empirical findings~\citep{golmant2018computational, shallue2018measuring}.
The shape of the curves is characteristically the same for different loss thresholds, though the critical batch size seems to increase for more difficult thresholds.

\subsection{Results of Optimal Learning Rate on NQM}
\begin{figure}[h]
\centering
\begin{subfigure}[t]{0.32\textwidth}
    \centering
    \includegraphics[height=1.2in]{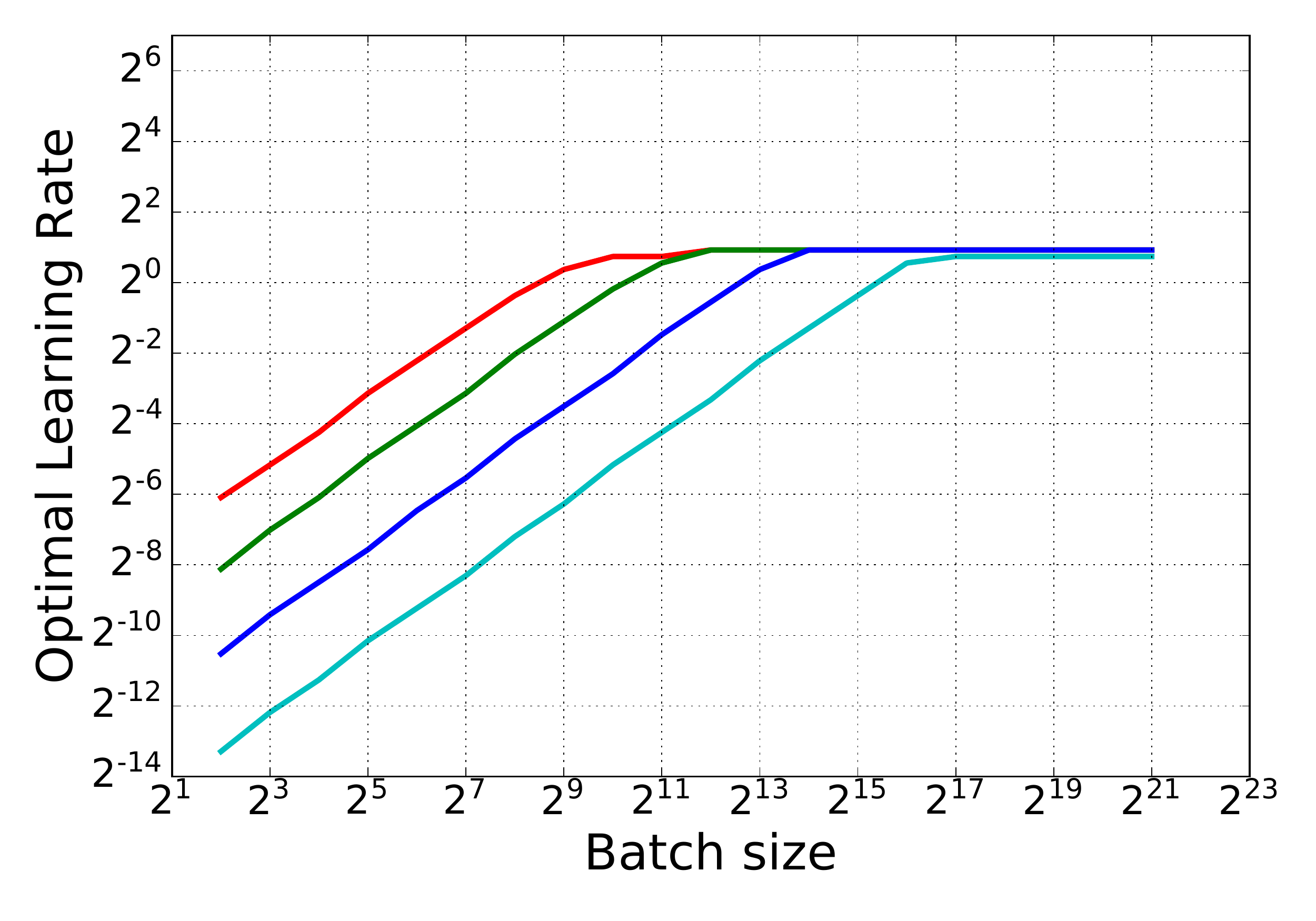}
    \vspace{-0.2cm}
    \caption{Without Momentum}
    \label{subfig:plain-sgd-optlr}
\end{subfigure}
\begin{subfigure}[t]{0.32\textwidth}
    \centering
    \includegraphics[height=1.2in]{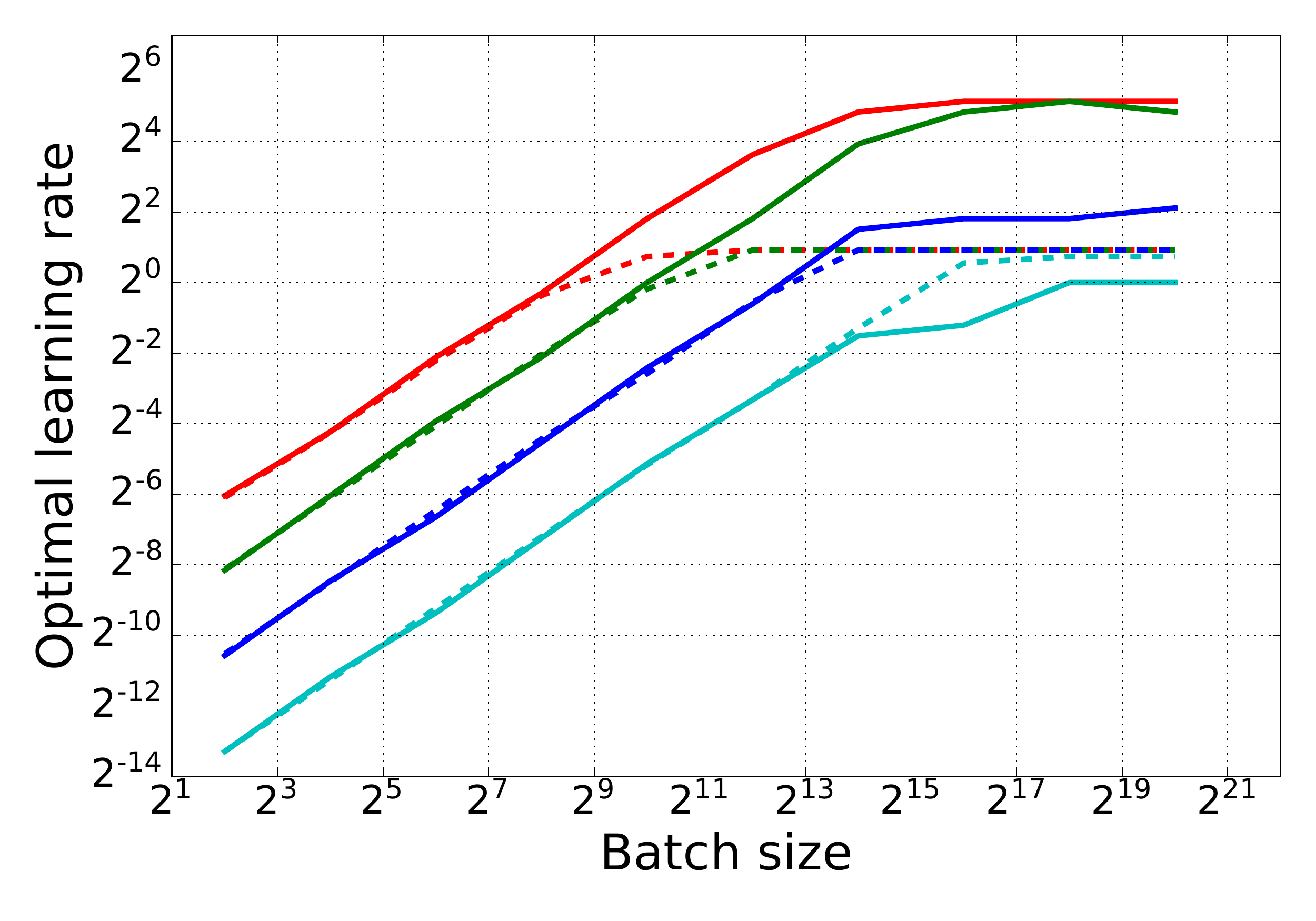}
    \vspace{-0.2cm}
    \caption{Fixed Momentum $0.9$}
\end{subfigure}
\begin{subfigure}[t]{0.32\textwidth}
    \centering
    \includegraphics[height=1.2in]{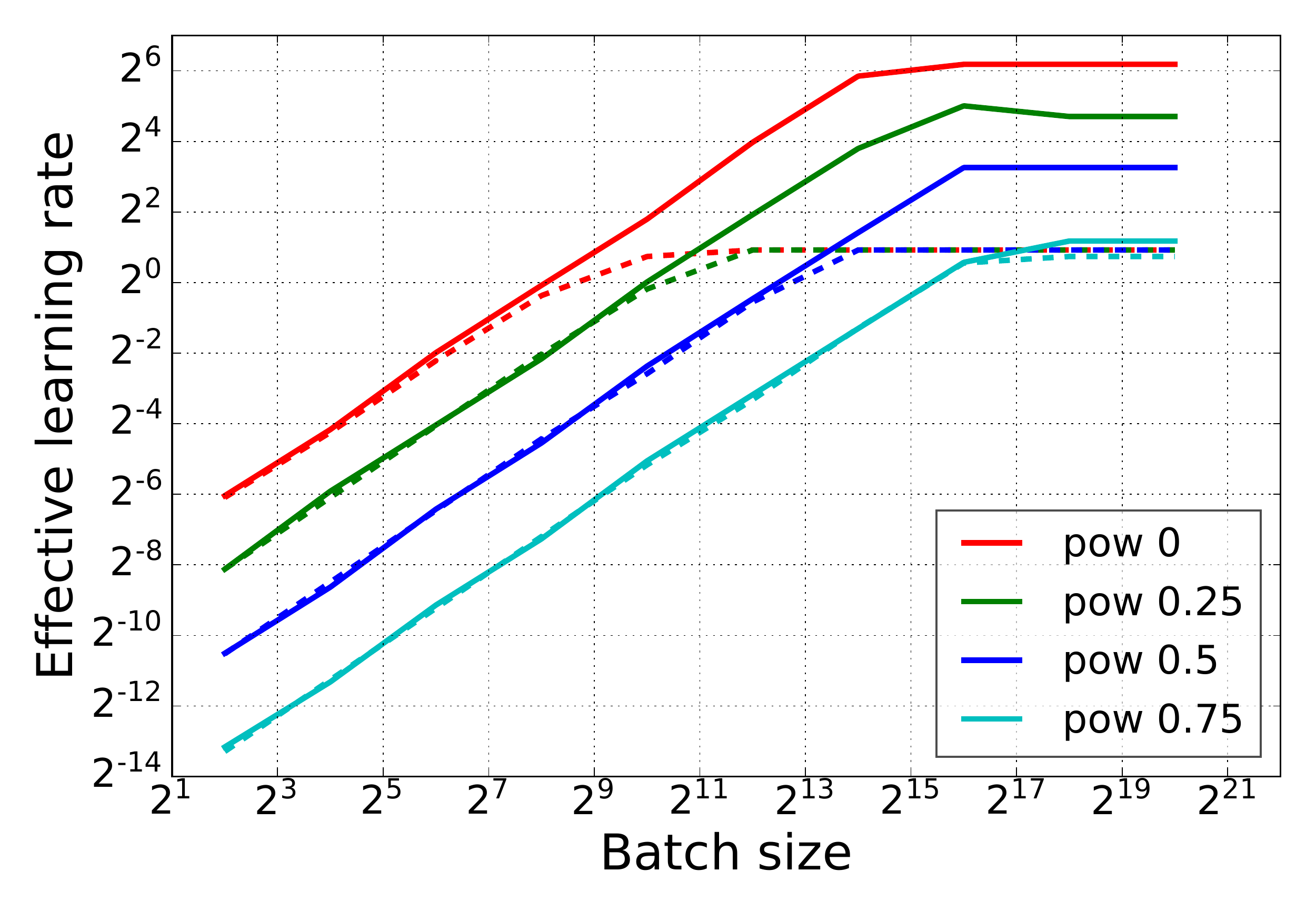}
    \vspace{-0.2cm}
    \caption{Tuned Momentum}
\end{subfigure}
\caption{\textbf{Optimal learning rate v.s. batch size for different preconditioning powers.} \textbf{(a)} When momentum is not used, the learning rate increases with batch size until it is limited by the maximum stable learning rate. Larger preconditioning powers reduce the optimal learning rate for the same batch size, thus extending the batch size where the learning rate levels off. \textbf{(b, c)} Fixed (0.9) and tuned momentum values. In (b) and (c), we plot the \emph{effective learning rate} for momentum SGD, defined as $\frac{\alpha}{1 - \beta}$. The dashed lines are the same plots from (a) for easier comparison.}
\vspace{-0.3cm}
\label{fig:optimal_lr} 
\end{figure}

\subsection{Final Learning Rate of Different Batch Sizes for PWC Learning Rate Scheme}
\label{app:final-lrate}
\begin{wrapfigure}[11]{R}{0.38\textwidth}
	\vspace{-0.6cm}
    \centering
    \includegraphics[width=1.9in]{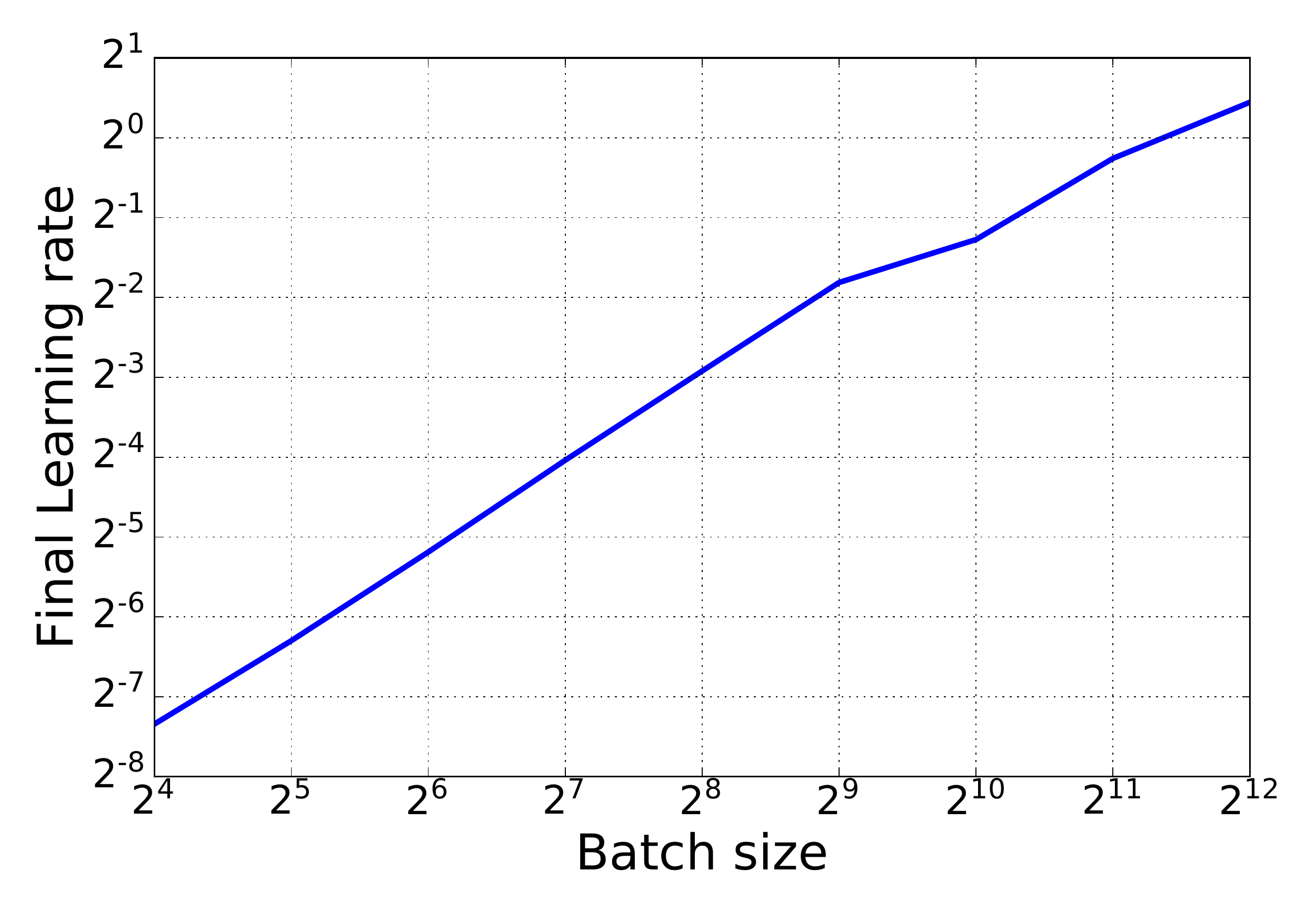}
    \vspace{-0.4cm}
    \caption{Final learning rate of the piecewise-constant learning rate scheme v.s. batch size.} %
    \label{fig:threshold}
    \vspace{-0.3cm}
    \label{fig:pwc_final_lrate}
\end{wrapfigure}
In Section~\ref{sec:lrate-scheme}, we study the piecewise constant learning rate scheme. The optimal scheme starts with a high learning rate which drops later in training (Figure~\ref{subfig:pwc_optlr}). Recall that for fixed learning rates, we observed that the optimal learning rate scaled linearly with the batch size for small batch sizes, but it is unclear whether there is a similar phenomenon for learning rate decay. In Figure~\ref{fig:pwc_final_lrate}, we plot the final learning rate as a function of batch size and show that it also scales linearly with batch size.

\section{More Details for Experiments}\label{app:experiments}

\subsection{Data Sets}
The data sets in Table \ref{tab:datasets_and_models} (MNIST, Fashion MNIST, CIFAR10, ImageNet and LM1B) are identical to those of~\citet{shallue2018measuring} (described in their Appendix A.1). For CIFAR10 we used data augmentation (including horizontal flip and random crop), but they did not.

\subsection{Model Details}\label{app:model-details}
This section provides details of models in Table \ref{tab:datasets_and_models}. The models are very similar (and some identical) to those used in~\citet{shallue2018measuring} (described in their Appendix B). Any modifications from them are highlighted in this section.

\textbf{Simple CNN} consists of 2 convolutional layers with max-pooling followed by 1 fully connected hidden layer. The convolutional layers use 5×5 filters with stride length 1, ``same'' padding \citep{goodfellow2016deep}, and ReLU activation function. Max pooling uses 2×2 windows with stride length 2. Unlike in~\citet{shallue2018measuring}, we did not use any dropout regularization (while they used dropout with probability 0.4 in the fully connected layer). We used 32 and 64 filters in the convolutional layers and 1,024 units in the fully connected layer. This corresponds to the ``base'' configuration in~\citet{shallue2018measuring}.

\textbf{ResNet8}~\citep{he2016deep} consists of 7 convolutional layers with residual connections followed by 1 fully connected hidden layer. We used the identical architecture as~\citet{shallue2018measuring}. In particular, we did not use batch normalization. The only difference is that we used data augmentation in our experiments.

\textbf{ResNet32}~\citep{he2016deep} consists of 31 convolutional layers with residual connections followed by 1 fully connected hidden layer (see Section 4.2 of~\citet{he2016deep}). We replaced batch
normalization~\citep{ioffe2015batch} with ghost batch normalization to keep the training objective fixed between batch sizes and to avoid possible negative effects from computing batch normalization statistics over a large number of examples~\citep{hoffer2017train}. We used a ghost batch size of 32 for all experiments. We also applied label smoothing~\citep{szegedy2016rethinking} to regularize the model at training time, which was helpful for larger batch sizes. We set the label smoothing parameter to 0.1 in all experiments. Instead of using weight decay, we applied channel-wise weight normalization by constraining the Frobenius norm of each convolutional channel to be exactly 1, which controls the effective learning rate~\citep{zhang2018three, van2017l2}.

\textbf{VGG11}~\citep{simonyan2014very} consists of 8 convolutional layers followed by 1 fully connected hidden layers. as in ResNet32, we used Ghost batch normalization, label smoothing, and channel-wise weight normalization.

\textbf{Transformer}~\citet{vaswani2017attention} is a self-attention model. We chose the Transformer model identical to the ``base'' model described in~\citet{vaswani2017attention}, except with only two hidden layers instead of six. This is identical to the ``Transformer Shallow'' model in~\citet{shallue2018measuring}.

\subsection{Learning Rate Schedules}

This section describes two learning rate schedules mentioned in Table \ref{tab:datasets_and_models}: constant schedule and linear decay schedule. Constant schedule simply keeps a fixed learning rate throughout training:
$$ \alpha(t) = \alpha_0, $$
where $t$ is the training step index. Linear decay schedule is
$$ \alpha(t) = \alpha_0 - (1 - \gamma) \frac{t}{T}, $$
where $\alpha_0$ is the initial learning rate, $\gamma$ is the rate of decay, and $T$ is the number of steps taken to reach the final learning rate. \citet{shallue2018measuring} experimented with various learning rate schedules and found that linear decay matched performance of the other schedules with fewer hyperparameters to tune. Therefore, we also chose the linear decay schedule, for which we tuned $\alpha_0$, $\gamma$ and $T$.

\subsection{Optimizer-Specific Hyperparamters}\label{app:optimizer-parameters}
For momentum SGD, we tuned the momentum $\beta$. For Adam, we tuned $\beta_1$, $\beta_2$, and $\epsilon$ (see \citet{kingma2014adam}). For K-FAC, we tuned damping and the trust region constraint (also known as the KL clipping term) for Transformer, keeping momentum $=0.9$ and the moving average parameter for damping $=0.99$; for all other models, we tuned all four parameters (see \citet{martens2015optimizing}).

\end{document}